\documentclass{article}

\PassOptionsToPackage{numbers,sort}{natbib}

\usepackage[main, final]{neurips_2026}

\usepackage{xcolor}
\definecolor{citecolor}{HTML}{0071bc}
\usepackage[utf8]{inputenc}
\usepackage[T1]{fontenc}
\usepackage[
  pagebackref=true,
  breaklinks=true,
  colorlinks,
  citecolor=citecolor,
  urlcolor=citecolor,
  linkcolor=black,
  bookmarks=false
]{hyperref}

\usepackage{url}
\usepackage{booktabs}
\usepackage{amsfonts}
\usepackage{nicefrac}
\usepackage{microtype}
\usepackage{algorithm}
\usepackage{algorithmic}
\usepackage{amsmath}
\usepackage{amssymb}
\usepackage{mathtools}
\usepackage{amsthm}
\usepackage[capitalize]{cleveref}
\usepackage{booktabs}
\usepackage{multirow}
\usepackage{graphicx}
\usepackage{adjustbox}
\usepackage{amsmath}
\usepackage{caption}
\usepackage{threeparttable}
\usepackage{siunitx}
\usepackage{enumitem}
\usepackage[font=small]{caption}
\usepackage{float}
\newcommand{\method}{\textbf{\textsc{ID-WM}}}
\newcommand{\superscript}[1]{\textsuperscript{\textnormal{#1}}}

\title{Hierarchical World Models with Implicit Dynamics}

\author{
    \textbf{Gaoyue Zhou\superscript{1,2}} \quad
    \textbf{Yvonne Wu\superscript{1}} \quad
    \textbf{Zichen Jeff Cui\superscript{1}}
    \\[4pt]
    \textbf{Nicolas Ballas\superscript{2}} \quad
    \textbf{Mido Assran\superscript{2}} \quad
    \textbf{Lerrel Pinto\superscript{1}} \quad
    \textbf{Yann LeCun\superscript{1,3}}
    \\[6pt]
    \superscript{1}Courant Institute, New York University \quad
    \superscript{2}Meta-FAIR \quad
    \superscript{3}AMI Labs
    \\[2pt]
    Correspondence: \texttt{gz2123@nyu.edu}
}

\begin{document}

\maketitle

\begin{abstract}
Planning with world models enables test-time adaptation for embodied agents, offering superior generalization over direct policy learning. However, long-horizon tasks remain hindered by cumulative errors and search complexity. Existing hierarchical methods often use rigid macro-actions or latent skills, leading to brittle representations and limited fine-grained control. We introduce \textbf{\method{} (Implicit Dynamics World Models)}, a hierarchical world model that decouples state-space reachability from local control. Our framework utilizes a high-level model to sample feasible transitions across the state manifold and a low-level world model as an inverse dynamics solver to ground these transitions into precise actions. This allows the planner to identify reachable subgoals without being constrained by low-level execution details. By prioritizing local physical feasibility over global behavioral patterns, \method{} synthesizes novel long-horizon plans from short segments, transcending the specific behavioral modes seen during training. We demonstrate its effectiveness across three navigation and manipulation suites, achieving a 38.5\% improvement over state-of-the-art policy and diffusion-based planners. Crucially, by leveraging this hierarchical structure to mitigate compounding errors, our approach yields a 135\% performance gain over single-level world model MPC, particularly in complex, unseen configurations for long-horizon tasks. Videos can be viewed at \url{https://id-wm.github.io}.

\end{abstract}

\section{Introduction}

World models (WM) ~\citep{truck-backer-upper, wm, lecun2022path} have emerged as a powerful paradigm for embodied decision-making~\citep{gaia1, dreamerv4, newt, Li2026CausalWM}. By learning predictive models of environment dynamics, they enable test-time reasoning and the synthesis of novel behaviors for unseen goals by searching in the action space~\citep{dinowm, vjepa2, nwm, dexwm}. However, planning with WMs faces two fundamental challenges: 1) search complexity grows exponentially with planning horizon, and 2) prediction errors accumulate over long rollouts. Therefore, most WM-based planners are restricted to short-horizon tasks~\citep{dinowm, vjepa2}, rely on simplified action abstractions~\citep{nwm}, or operate in high-level spaces such as language, sacrificing fine-grained control~\citep{vlwm}.

Hierarchical learning offers a principled solution to this scalability challenge by decomposing long-horizon problems into manageable abstractions~\citep{schmidhuber1991learning, Sutton1999BetweenMA, Barto2003RecentAI}. It has been extensively studied in Reinforcement Learning (RL)~\citep{FeUdal, hiro, hiql, Pertsch2020AcceleratingRL}, where hierarchy mitigates the credit assignment problem in long trajectories, and in imitation learning~\citep{Gupta2019RelayPL,saycan}, where it enables reusable behaviors. More recently, hierarchical structure has been introduced into diffusion-based planners to avoid directly modeling entire trajectories~\citep{Li2023HierarchicalDF, hdiffuser, Hao2025CHDCH}, improving scalability and generalization beyond fixed-horizon training data.

How can we enable the same capabilities for task-agnostic world models, enabling long-horizon zero-shot planning for unseen goals? A direct approach would be to simply increase the prediction horizon or action abstraction (e.g., frame-skipping)~\citep{Zhang2026HierarchicalPW}. However, previous attempts at hierarchical world models typically depend on task-specific priors to function effectively, relying on pre-trained skills and low-level policies~\citep{lsp, Pertsch2020AcceleratingRL} or rewards and value functions to verify subgoal feasibility~\citep{Gumbsch2024LearningHW, Nasiriany2019PlanningWG}. This dependency arises because planning is fundamentally adversarial: without these anchoring constraints, a powerful optimizer will inevitably exploit deficiencies in a learned world model, discovering "wormholes" or unrealistic transitions~\cite{Chua2018DeepRL, psenka2026parallelstochasticgradientbasedplanning}. To make hierarchical world models suitable for planning, we must construct a latent space that both condenses trajectory information to minimize search complexity, and remains inherently well-behaved to prevent the planner from exploiting unrealistic transitions.

\begin{figure*}[t!]
    \centering
    \includegraphics[width=\textwidth]{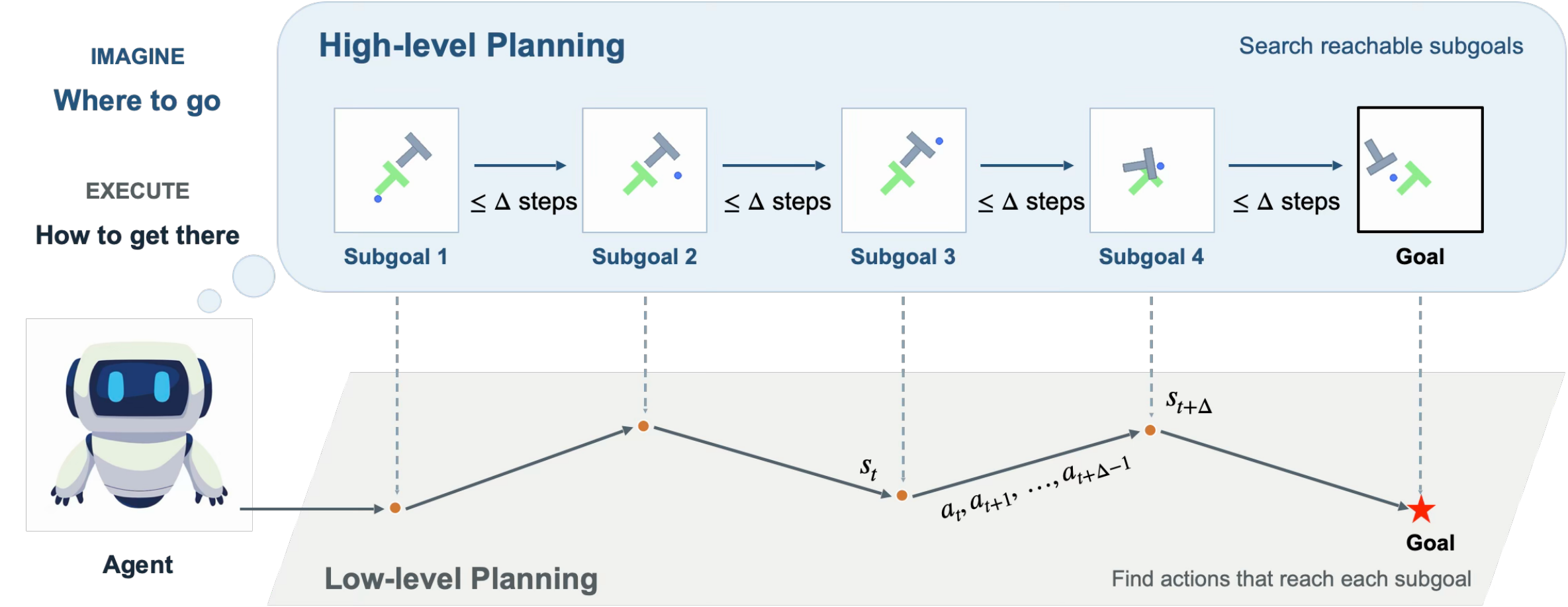}
    \caption{\textbf{Overview of \method{}.} Our hierarchical planning framework decouples state-space reachability from local control, eliminating the reliance on task-specific value functions or pre-trained policies. Planning proceeds in two interleaved stages: (1) high-level planning, which discovers a sequence of temporally abstract subgoals spaced $\Delta$ steps apart, and (2) low-level control, which leverages a world model as an inverse dynamics solver to extract executable actions between each subgoal.}
    \vspace{-3mm}
    \label{fig:teaser}
\end{figure*}

To this end, we introduce \textbf{\method{}}, a hierarchical world model with implicit dynamics grounded in the insight that effective planning requires decoupling topological reachability from execution details (\Cref{fig:teaser}). \method{} consists of a high-level diffusion-based world model that only implicitly models the dynamics by outputting states that are within a fixed distance away from the current state, functioning as a \emph{neighboring state generator}, and a deterministic low-level model conditioned on both current state and action, acting as an \emph{inverse dynamics solver}. Conditioned on the current state and a Gaussian noise, the high-level model leverages the generative power of diffusion to map a tractable unit Gaussian latent space to the manifold of feasible neighboring states. This formulation allows us to treat the noise as an implicit high-level action, enabling the planner to search over a well-behaved distribution instead of complex, discontinuous skill spaces to reason purely about "where to go," while the low-level model recovers the precise executable actions for "how to get there."

We summarize our contributions as follows:
\begin{itemize}
    \item We introduce \textbf{\method{}}, a hierarchical world model that decouples reachability from control, eliminating the need for task-specific value functions or pre-trained policies.
    \item We propose a corresponding hierarchical planning algorithm that bridges high-level abstraction with local dynamics, enabling robust, long-horizon decision-making.
    \item Across three navigation and manipulation environment suites, \method{} shows 38.5\% improvement for reaching arbitrary goals against powerful hierarchical offline reinforcement learning, imitation learning, and diffusion planning methods, and crucially, a 135\% improvement over single-level world model and planning regimes.
\end{itemize}

\section{Related Work}

\subsection{Hierarchical Learning}

Hierarchical learning has long been a foundational pursuit in artificial intelligence, decomposing long-horizon problems into manageable abstractions~\citep{schmidhuber1991learning, Sutton1999BetweenMA, Barto2003RecentAI}. This paradigm spans policy architectures~\citep{Chung2016HierarchicalMR}, reinforcement learning~\citep{FeUdal, hiql, Haramati2026HierarchicalER, Gumbsch2024LearningHW}, imitation learning~\citep{Gupta2019RelayPL}, and planning~\citep{Nair2019HierarchicalFS, Nasiriany2019PlanningWG}. Existing work focuses on both \textit{behavioral} (skills/options)~\citep{Sutton1999BetweenMA, lsp, Pertsch2020AcceleratingRL, SkiMo, Dalal2021AcceleratingRR, Zhang2023LeveragingJM, puppeteer} and \textit{subgoal-centric} approaches~\citep{HiGoC, Nasiriany2019PlanningWG, hiro}. However, these methods typically rely on coupled priors, such as rewards, learned value functions or pre-trained low-level policies~\citep{Nasiriany2019PlanningWG, Haramati2026HierarchicalER}~to ensure feasibility, limiting generalization beyond the training distribution. In contrast, we eliminate this dependency, enabling robust hierarchical reasoning purely through implicit dynamics even when reliable value functions and policies are absent.

\subsection{World Models for Embodied Planning}

World models~\citep{truck-backer-upper, wm, lecun2022path} have emerged as a foundational paradigm for embodied decision-making, enabling agents to reason about the future consequences of their actions. Recently, the field has shifted towards developing general-purpose world models~\citep{genie, dinowm, nwm, sora2, veo3, dreamerv4} capable of capturing broad environmental dynamics, diverging from traditional model-based reinforcement learning (MBRL) where learning task-specific reward or value functions is paramount~\citep{dreamer, dreamerv2, dreamerv3, tdmpc, tdmpc2, transformer100k, iris}. Once these dynamics are learned, inference-time planning algorithms like the Cross-Entropy Method (CEM)~\citep{cem} or gradient descent—are employed to derive behaviors zero-shot~\citep{dinowm, vjepa2, pldm, dexwm}. However, the effectiveness of these approaches is often constrained to short horizons due to the computational complexity of search and the accumulation of prediction errors over time. While various strategies have been proposed to mitigate these issues, ranging from robust gradient-based optimization~\citep{robustwmplanning, psenka2026parallelstochasticgradientbasedplanning} and efficient sampling routines~\citep{Bharadhwaj2020ModelPredictiveCV} to temporal approximations for longer action sequences~\citep{nwm}, they remain fundamentally flat in their temporal reasoning. Consequently, they struggle with long-horizon task solving, remaining susceptible to local minima and limited in their scalability to complex behaviors. Concurrently, HWM~\citep{Zhang2026HierarchicalPW} builds a hierarchical latent world model by increasing the high-level frameskip and learning macro-actions; in contrast, \method{} searches over the diffusion noise space, which is smooth and well-conditioned by construction.

\subsection{Diffusion Models and Planning}

Diffusion probabilistic models~\citep{diffusion, ddpm, ddim} excel at capturing complex, multimodal distributions via iterative denoising. Their efficacy is proven across high-dimensional modalities, including pixels~\citep{betker2023improving, Saharia2022PhotorealisticTD, sora2, veo3}, language~\citep{Li2022DiffusionLMIC}, and robot actions~\citep{diffusionpolicy, ditpolicy}. Recently, diffusion models have been proposed as zero-shot planners for embodied tasks~\citep{diffuser, Ajay2022IsCG, Lu2025WhatMA, avdc}. These methods model entire trajectories directly, limiting generalization to out-of-distribution initial-goal pairs and constraining planning horizons to those seen during training. While hierarchical formulations mitigate these constraints, they rely on task-specific value functions~\citep{Li2023HierarchicalDF, Haramati2026HierarchicalER, hdiffuser}, successful task trajectories~\citep{hdflow}, or assumed access to low-level primitives~\citep{Hao2025CHDCH}. In contrast, we redefine diffusion as a \textit{local generator} rather than a \textit{global planner}. By delegating long-horizon reasoning to search, our method composes local jumps into novel trajectories without requiring task-specific priors.

\section{Hierarchical World Models with Implicit Dynamics} 

\begin{figure*}[t!]
    \centering
    \includegraphics[width=1\textwidth]{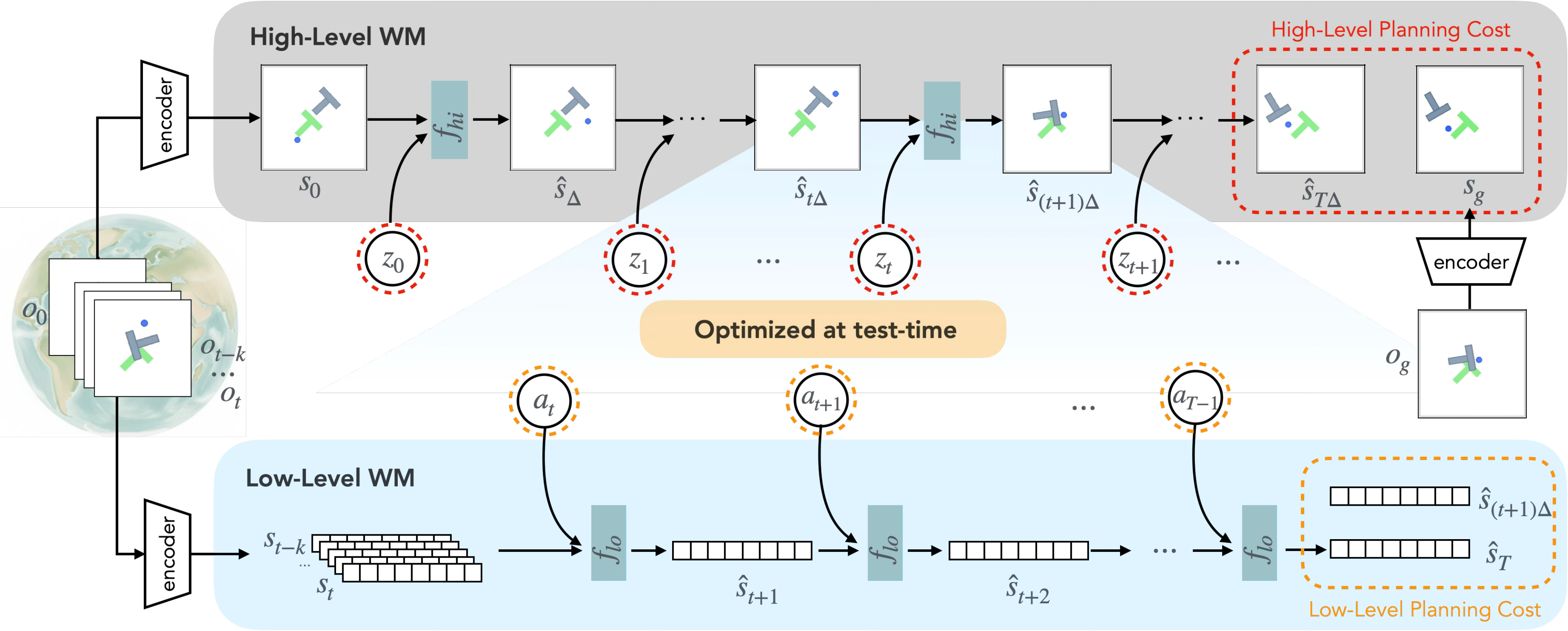}
    \caption{Architecture of \method{}. Given an initial observation $o_0$ and a goal $o_g$, we optimize for both the sequence of subgoals $s_0, \dots, s_{T\Delta}$ and the executable actions $a_t$ in an interleaved fashion. Here, $f_{hi}$ denotes the diffusion-based high-level world model and $f_{lo}$ denotes the deterministic, transformer-based low-level world model, both of which predict in latent space.}
    \label{fig:arch}
\end{figure*}

\textbf{Overview and Problem Formulation.}
We consider a goal-conditioned planning problem within a Markov Decision Process (MDP) defined by the tuple $\mathcal{M} = (\mathcal{S}, \mathcal{A}, \mathcal{P})$, where $\mathcal{S}$ is the continuous state space, $\mathcal{A}$ is the action space, and $\mathcal{P}(s_{t+1}|s_{\le t}, a_{\le t})$ represents the transition dynamics.

Following prior world modeling 
methods~\citep{dinowm, nwm, dexwm}, we learn environment dynamics from a task-agnostic offline dataset without any reward labeling via a world model, and plan to reach arbitrary unseen goals at test time. More specifically, an agent is provided with an 
initial state $o_0$ and a goal state $o_g$ at test time. The objective is to execute a sequence 
of actions $(a_0, a_1, \dots, a_{T-1})$ such that the distance between the agent's 
final state $o_T$ and the goal $o_g$ is minimized. Unlike prior work, we do not 
assume a fixed task horizon $T$~\cite{dinowm}, focusing instead on enabling the agent to 
perform autonomous long-horizon reasoning and planning.

\subsection{Hierarchical World Models with Implicit Dynamics}

The core component of \method{} is a hierarchical world model comprising a high-level diffusion model and a low-level inverse dynamics model. As illustrated in Figure~\ref{fig:arch}, the high-level model operates as a \textit{reachable state generator}, predicting a distribution of feasible future states conditioned on the current state. The low-level model acts as an \textit{inverse dynamics solver}, grounding these predicted transitions into executable actions.
\begin{align*}
\text{High-level WM:} & \quad f_{hi}(s_{t+\Delta} \mid s_{t}, \Delta) \\
\text{Low-level WM:} & \quad f_{lo}(s_{t+1} \mid s_{t-k:t}, a_{t-k:t})
\end{align*}
Specifically, the high-level world model conditions on the current state $s_t$ and a temporal radius $\Delta$, generating future states that fall within a $\Delta$-step horizon. The low-level world model adopts the formulation from~\citet{dinowm}, a transformer-based world model conditioning on a historical context of $k$ past states and actions to infer immediate next steps.

Both levels of the hierarchy share a unified structure, consisting of the following components:
\begin{align*}
    \text{Observation model:} & \quad s_t \sim \text{enc}_\theta(s_t \mid o_t) \\
    \text{Transition model:} & \quad s_{t+1} \sim p_\theta(s_{t+1} \mid s_{\leq t}, c) \\
    \text{Decoder model:} & \quad \hat{o}_t \sim q_\theta(o_t \mid s_t) \\
\end{align*}
where $c$ is the context that the transition model takes in to make predictions. For $f_{hi}$, $c$ is the temporal distance $\Delta$, instructing how far away the subgoals should be. For $f_{lo}$, $c$ is the sequence of executable actions the agent has taken up to time $t$ (i.e., $a_{\leq t}$).

\subsubsection{High-level Implicit Dynamics Model}
\label{sec:diffusion_model}

To model the manifold of reachable states without restricting the agent to a fixed set of discrete skills, we formulate the high-level world model as a conditional diffusion probabilistic model (DDPM)~\citep{ddpm}. 
Let $x_0$ represent the target future state $s_{t+\Delta}$ (at a fixed temporal offset $\Delta$). The forward diffusion process $q(x_j | x_{j-1})$ gradually adds Gaussian noise to the data according to a variance schedule $\beta_1, \dots, \beta_K$:
\begin{equation}
    q(x_j | x_{j-1}) = \mathcal{N}(x_j; \sqrt{1 - \beta_j} x_{j-1}, \beta_j \mathbf{I})
\end{equation}
This allows us to sample $x_j$ at any arbitrary diffusion step $j$ in closed form. We learn a conditional denoising network $\epsilon_\theta(x_j, j, s_t)$ to reverse this process, where $s_t$ is the current state to condition the generation. The training objective is to predict the noise $\epsilon$ added to the clean data $x_0$:
\begin{equation}
    \mathcal{L}_{\text{diff}} = \mathbb{E}_{x_0, \epsilon \sim \mathcal{N}(0, I), j} \left[ \| \epsilon - \epsilon_\theta(x_j, j, s_t) \|^2 \right]
\end{equation}

\textbf{Sampling as Implicit Dynamics (DDIM).}
To represent the high-level decision at environment time $t$, we let $z_t \in \mathbb{R}^d$ denote the initial noise vector $x_K$ used for sampling the next subgoal. During inference, we employ Denoising Diffusion Implicit Models (DDIM)~\citep{ddim} to deterministically map this noise vector $z_t \sim \mathcal{N}(0, \mathbf{I})$ to a generated subgoal $\hat{s}_{t+\Delta}$. The deterministic reverse step from diffusion step $j$ to $j-1$ is given by:
\begin{equation}
\begin{split}
    x_{j-1} &= \sqrt{\bar{\alpha}_{j-1}} \underbrace{\left( \frac{x_j - \sqrt{1 - \bar{\alpha}_j} \epsilon_\theta(x_j, j, s_t)}{\sqrt{\bar{\alpha}_j}} \right)}_{\text{predicted } x_0} \\
    &\quad + \sqrt{1 - \bar{\alpha}_{j-1}} \epsilon_\theta(x_j, j, s_t)
\end{split}
\end{equation}
where $\alpha_j = 1 - \beta_j$ and $\bar{\alpha}_j = \prod_{i=1}^j \alpha_i$. This formulation effectively parameterizes the high-level transition dynamics as a deterministic function $\mathcal{G}_\theta(s_t, z_t)$.

\textbf{Implicit High-Level Action Space.}
A key insight of our approach is to treat the initial noise vector $z_t$ as an \emph{implicit high-level action}. Unlike discrete skill indices or learned latent codes which may suffer from mode collapse or topological discontinuities, illustrated in ~\cref{app:vs}, the noise space $\mathcal{Z} = \mathbb{R}^d$ is inherently smooth and well-behaved (isotropic Gaussian).
By optimizing over $z_t$, the planner searches over the standard Gaussian latent space, where any sample $z$ maps to a physically valid reachable state on the data manifold via the deterministic DDIM mapping.

\subsubsection{Low-level World Model}
\label{sec:low_level_wm}

The high-level World Model (WM) provides long-horizon guidance by identifying target states, yet it lacks the granular transition dynamics necessary to prescribe executable actions. To this end, we employ a transformer-based deterministic low-level WM following \citet{dinowm}, which functions as an inverse dynamics model through planning.

Given a history of states $s_{t-k:t}$ and actions $a_{t-k:t}$, the low-level WM predicts the next state $\hat{s}_{t+1} = f_{lo}(s_{t-k:t}, a_{t-k:t})$. The model is trained via supervised learning to minimize the prediction error:
\begin{equation}
    \mathcal{L} = \|\hat{s}_{t+1} - s_{t+1}\|_2
\end{equation}
Consistent with \citet{dinowm}, we utilize an MLP action encoder to process raw action sequences before they are input into the transformer backbone.

At inference, we leverage this model for inverse dynamics through planning. Given a target state $s^*$, we search for an optimal action sequence $(a_0, \dots, a_{H_{lo}-1})$ over a low-level planning horizon $H_{lo}$ such that the planning cost $\|\hat{s}_{H_{lo}} - s^*\|_2$ is minimized. We detail the full hierarchical planning procedure in Section~\ref{sec:planning}.

\subsection{Hierarchical Planning with \method{}}
\label{sec:planning}

At inference, given an initial state $o_0$ and a goal state $o_g$, we employ hierarchical planning to determine a sequence of actions that reaches $o_g$. We formulate this as a Model Predictive Control (MPC) problem, where each iteration consists of two phases: subgoal generation and subgoal realization. The overall hierarchical planning procedure is detailed in ~\Cref{alg:reduced}. We also include a full version in~\Cref{alg:verbose}. Throughout, $\Delta$ denotes the temporal radius of each subgoal (the maximum number of environment steps between consecutive subgoals), $H_{hi}$ the number of subgoals per high-level plan, and $H_{lo}$ the low-level planning horizon used to reach each subgoal.

\textbf{Subgoal Search via High-level WM.} In the first phase, we identify a sequence of latent subgoals $s^*_0, \dots, s^*_{H_{hi}}$ such that $s^*_0 = s_0$ and $s^*_{H_{hi}} = s_g$. To ensure that each subgoal remains within the reachable manifold of the low-level WM given its search budget, we constrain the distance between consecutive subgoals $s^*_i, s^*_{i+1}$ to be no more than $\Delta$ environment steps. We optimize the latent noise variables $z_0, \dots, z_{H_{hi}-1}$ using the Cross-Entropy Method (CEM) to minimize the following planning cost:
\begin{equation}
    \mathcal{L} = \|\hat{s}_{H_{hi}} - s_g\|_2 + \alpha \mathcal{L}_{reg}(Z)
\end{equation}
where $\mathcal{L}_{reg}$ is a regularization term designed to keep the sampled noise within the support of the standard Gaussian distribution. This prevents the planner from exploiting adversarial regions of the latent space that would result in out-of-distribution transitions during DDIM sampling.

\textbf{Subgoal Realization via Low-level WM.} We similarly employ the CEM optimizer to identify the executable action sequences required to reach each intermediate subgoal. The agent plans and executes actions to reach the immediate subgoal, proceeding sequentially through the high-level trajectory. Upon completing the sequence for all subgoals generated by the high-level WM, the agent initiates a new MPC iteration to replan the high-level trajectory from its current state.

\textbf{Latent Noise Regularization.}
While the latent space is theoretically structured as a unit Gaussian, unconstrained planning can favor adversarial latents that produce unrealistic state transitions. To mitigate this, we incorporate a regularization term in the high-level planning cost:
\begin{equation}
    \mathcal{L}_{reg} = \mathbb{E}_{i} \left[ \left( \max(0, |z_{i}| - \tau) \right)^p \right]
\end{equation}
This term penalizes latent vectors whose magnitude exceeds a threshold $\tau$, ensuring the optimization stays within high-density regions of the prior. We examine the effect of noise regularization in~\cref{appendix:ab_plan_param}.

\begin{algorithm}[h]
\caption{Hierarchical Planning with \method{}}
\label{alg:reduced}
\begin{algorithmic}[1]
\REQUIRE Initial state $s_0$, goal state $s_g$, high/low models $f_{hi}, f_{lo}$, planning costs $C_{hi}, C_{lo}$, planning horizons $H_{hi}, H_{lo}$.
\STATE $s_t \gets s_0$
\WHILE{goal $s_g$ not reached}
    \STATE \textcolor{teal}{\ttfamily{\# High-Level Plan}}
    \STATE $(\hat{s}_1, \dots, \hat{s}_{H_{hi}}) \gets \text{Planner}(s_t, s_g, f_{hi}, C_{hi})$
    
    \STATE \textcolor{teal}{\ttfamily{\# Low-Level Plan and Execution}}
    \FOR{$i = 1$ to $H_{hi}$}
        \STATE $(a_1, \dots, a_{H_{lo}}) \gets \text{Planner}(s_t, \hat{s}_i, f_{lo}, C_{lo})$
        \STATE Execute actions $(a_1, \dots, a_{H_{lo}})$ in the environment
        \STATE $s_t \gets$ observe new state
    \ENDFOR
\ENDWHILE
\end{algorithmic}
\end{algorithm}

\section{Experiments}

Our experiments are designed to answer the following key questions:
\begin{itemize}
    \item \textbf{Subgoal Generation:} Can \method{} reliably generate diverse and feasible reachable states?
    \item \textbf{Generalization:} Can \method{} solve long-horizon planning tasks to reach arbitrary, zero-shot goals at test time?
    \item \textbf{Scaling:} How does the performance of \method{} scale across environments of increasing complexity?
    \item \textbf{Visual Planning:} Is \method{} effective for long-horizon planning directly from high-dimensional visual observations?
\end{itemize}

\subsection{Environments}

\begin{figure}[h]
    \centering
    \includegraphics[width=0.9\columnwidth]{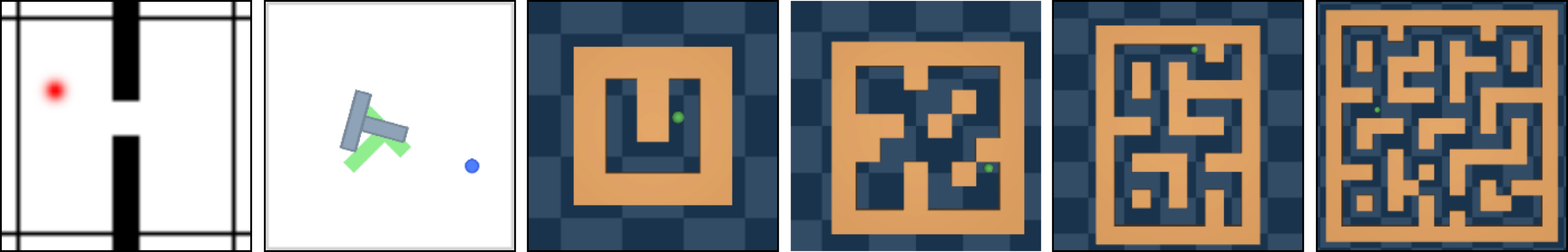}
    \caption{We evaluate \method{} on three environment suites (Wall, PushT, Maze); from left to right: Wall, PushT, and the four maze sizes (U-, Medium, Large, Giant).}
    \label{fig:envs}
\end{figure}

We evaluate~\method{} on three navigation and manipulation environment suites as shown in Figure~\ref{fig:envs}. The environments are Wall, PushT, and Maze, with task horizon ranging from hundreds to thousands of environment steps. A full description of the environments is detailed in~\Cref{appendix:envs}.

\subsection{Baselines}

We compare \method{} with five state-of-the-art offline learning methods including imitation learning (GCBC), offline RL (HIQL), diffusion planners (Diffuser, Hierarchical Diffuser), and planning with world models (Single-level WM with MPC). See~\Cref{appendix:baselines} for full details. 

\subsection{Generating Neighboring States with \method{}}
\label{sec:neighboring}

We first investigate the extent to which each level of \method{} accurately captures the underlying data distribution and predicts future states across varying temporal granularities. Specifically, for the high-level WM, we evaluate whether sampling from the latent prior yields diverse and reachable future states.

\begin{figure}[h]
    \centering
    \includegraphics[width=\columnwidth]{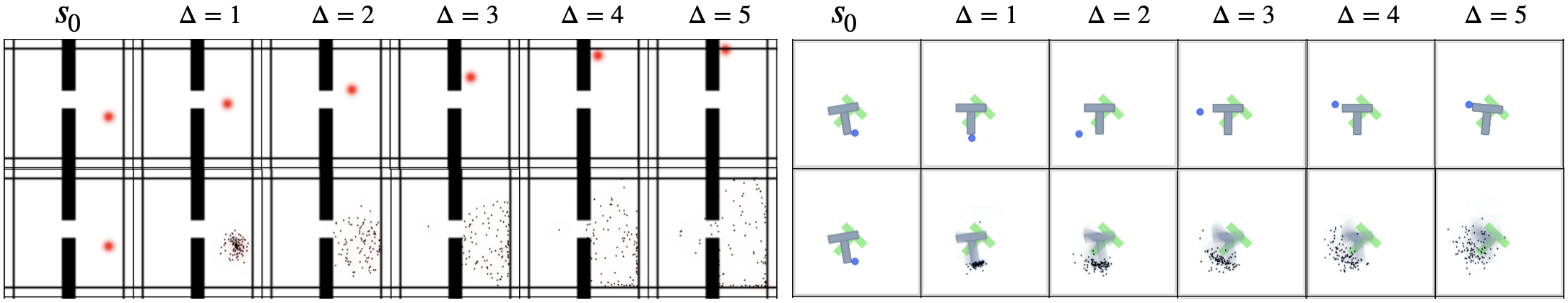}
    \caption{Diverse neighboring states generated with the high-level WM of \method{}. }
    \label{fig:next_states}
\end{figure}

In \Cref{fig:next_states}, we sample 100 latent noise vectors $z \sim \mathcal{N}(0, I)$ and visualize the resulting generated states $\hat{s}_\Delta = f_{hi}(s_0, z, \Delta)$ reachable within a $\Delta$-step horizon. For each $\Delta$, we overlay all 100 generated samples to visualize the predicted state distribution, explicitly marking the agent's position with a black dot for clarity.

As illustrated in \Cref{fig:next_states}, the high-level WM generates diverse and physically reachable states for both the Wall and PushT environments. In the Wall environment, the dispersion of agent positions increases with $\Delta$, reflecting a larger reachable radius as the temporal budget grows. Similarly, in the PushT environment, we observe an increasingly diffused distribution of both the agent and the T-block. The smooth coverage of object configurations demonstrates that the model successfully captures the complex, multi-modal interactions between the agent and the object. This high-fidelity distribution modeling serves as a necessary prerequisite for planning, enabling the system to generate diverse trajectories and generalize effectively to unseen goals in downstream tasks.

\subsection{Hierarchical Planning with~\method{}}

Can \method{} solve long-horizon planning tasks to reach arbitrary, zero-shot
goals at test time? To this end, we conduct hierarchical planning on three environment suites.

For all environments and baselines, we randomly sample 50 pairs of initial and goal states, with sampling method detailed in ~\Cref{sec:task_define}, and report the average performance across these instances. For the PushT task, we report two metrics: the success rate and the final coverage of the T-block relative to the target configuration. A trial is considered successful only if both the agent and the T-block reach their respective goal positions. In contrast, the final coverage metric provides a more granular measure of performance by focusing specifically on the object’s alignment. To further evaluate generalization capabilities, we test PushT under two settings: a standard regime and a data-augmented regime featuring random spatial transformations and rotations. All experiments are conducted with a maximum horizon of 2000 environment steps. For all experiments, \method{} uses $\Delta = 5$ (25 environment steps with frameskip) and $H_{hi} = 5$ subgoals per high-level plan, without per-task tuning; the planner replans after each executed plan (up to 15 replans), so no prior knowledge of the goal distance is required. We ablate $\Delta$ and the number of subgoals in \Cref{appendix:delta_ablation}.

\newcommand{\msd}[2]{$#1$\,{\scriptsize$\pm#2$}}
\newcommand{\bmsd}[2]{$\mathbf{#1}$\,{\scriptsize$\mathbf{\pm#2}$}}
\begin{table}[t]
\centering
\caption{Performance comparison across environments. We report success rate (SR) and, for PushT, final coverage (Cov), as mean $\pm$ std over three seeds. PushT is evaluated with a fixed goal (Fix) and random goals (Rand.). Best results per row are in bold.}
\label{tab:results}
\vspace{0.5em}
\small
\setlength{\tabcolsep}{4pt}
\begin{adjustbox}{max width=\columnwidth}
\begin{tabular}{@{}llcccccc@{}}
\toprule
\textbf{Environment} & \textbf{Metric} & \textbf{HIQL} & \textbf{GCBC} & \textbf{Diffuser} & \textbf{H-Diffuser} & \textbf{SL-WM} & \textbf{Ours} \\
\midrule
Medium Maze & SR & \msd{0.83}{0.01} & \msd{0.80}{0.00} & \msd{0.80}{0.13} & \msd{0.92}{0.03} & \msd{0.97}{0.01} & \bmsd{0.99}{0.01} \\
\midrule
Wall & SR & \msd{0.62}{0.12} & \bmsd{1.00}{0.00} & \msd{0.07}{0.08} & \msd{0.10}{0.02} & \msd{0.88}{0.02} & \bmsd{1.00}{0.00} \\
\midrule
\multirow{2}{*}{PushT (Fix)}
 & SR  & \msd{0.45}{0.07} & \bmsd{1.00}{0.00} & \msd{0.00}{0.00} & \msd{0.01}{0.01} & \msd{0.19}{0.02} & \bmsd{1.00}{0.00} \\
 & Cov & \msd{0.50}{0.05} & \bmsd{0.85}{0.04} & \msd{0.10}{0.01} & \msd{0.35}{0.09} & \msd{0.34}{0.03} & \bmsd{0.85}{0.01} \\
\midrule
\multirow{2}{*}{PushT (Rand.)}
 & SR  & \msd{0.02}{0.02} & \msd{0.21}{0.01} & \msd{0.00}{0.00} & \msd{0.01}{0.01} & \msd{0.21}{0.04} & \bmsd{0.77}{0.01} \\
 & Cov & \msd{0.22}{0.03} & \msd{0.31}{0.01} & \msd{0.05}{0.01} & \msd{0.20}{0.08} & \msd{0.27}{0.02} & \bmsd{0.58}{0.01} \\
\midrule
\multirow{2}{*}{PushT-Aug (Fix)}
 & SR  & \msd{0.87}{0.01} & \bmsd{1.00}{0.00} & \msd{0.01}{0.02} & \msd{0.55}{0.14} & \msd{0.10}{0.04} & \bmsd{1.00}{0.00} \\
 & Cov & \msd{0.81}{0.05} & \msd{0.83}{0.01} & \msd{0.31}{0.03} & \msd{0.74}{0.16} & \msd{0.43}{0.01} & \bmsd{0.84}{0.01} \\
\midrule
\multirow{2}{*}{PushT-Aug (Rand.)}
 & SR  & \msd{0.32}{0.06} & \msd{0.15}{0.01} & \msd{0.01}{0.01} & \msd{0.27}{0.03} & \msd{0.10}{0.04} & \bmsd{1.00}{0.01} \\
 & Cov & \msd{0.59}{0.03} & \msd{0.25}{0.01} & \msd{0.14}{0.01} & \msd{0.72}{0.19} & \msd{0.17}{0.01} & \bmsd{0.77}{0.01} \\
\bottomrule
\end{tabular}
\end{adjustbox}
\end{table}

\begin{figure}[t!]
    \centering
    \includegraphics[width=1\textwidth]{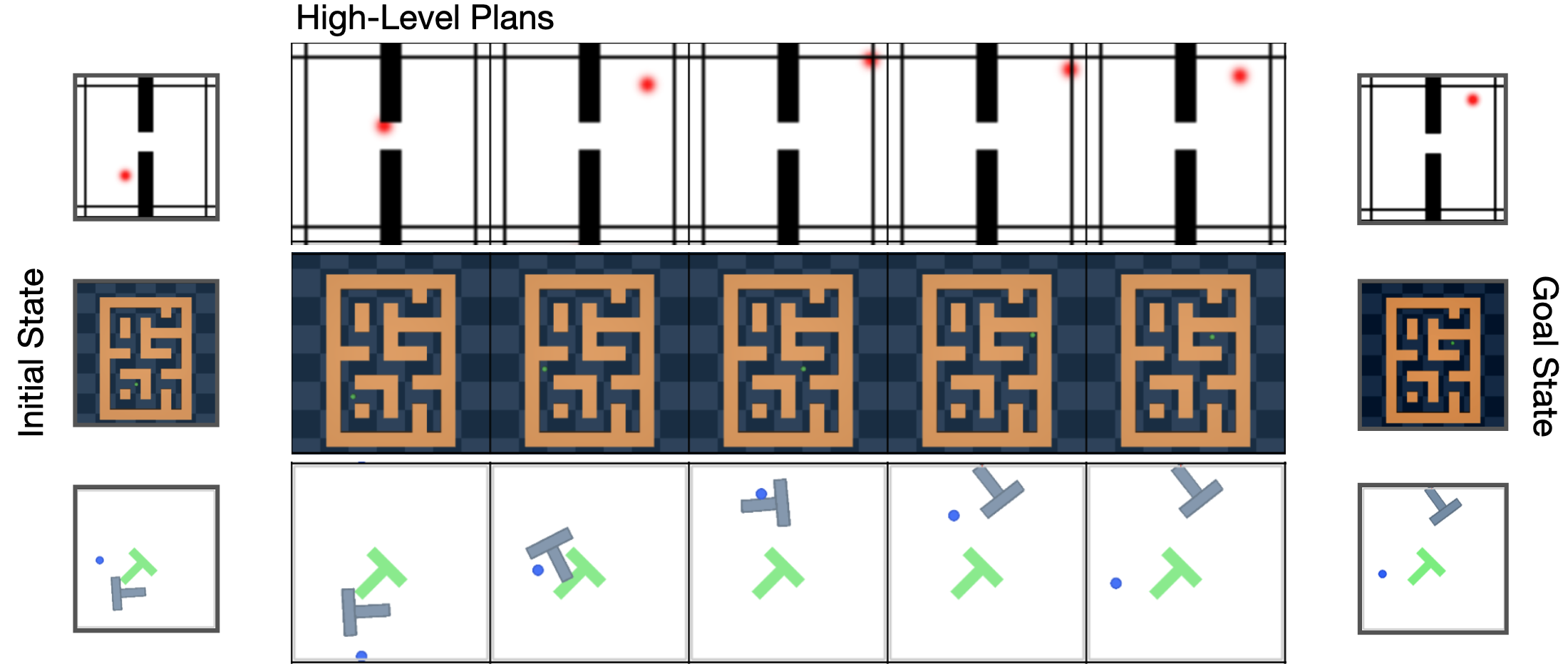}
    \caption{High-level subgoals planned by \method{}. The planned sequence successfully guides the agent to the target goal while ensuring that consecutive subgoals satisfy local reachability constraints.}
    \label{fig:subgoals}
\end{figure}

In \Cref{tab:results}, \method{} outperforms all baselines, achieving a 38.5\% overall improvement and 135\% gain over single-level WM with MPC (SL-WM). The gap is most evident in PushT with random goals, where \method{} is the only approach exceeding a 30\% success rate without augmentation. While SL-WM performs well on shorter horizons (Maze, Wall), it fails on PushT's long-horizon requirements (often $>100$ steps) due to exponential search complexity and compounding prediction errors. Unlike the 25-step benchmark in \citet{dinowm}, our results show that standard WM planning cannot solve these extended tasks, a limitation our hierarchical approach effectively addresses. We visualize the resulting high-level plans in \Cref{fig:subgoals} and examine subgoal quality and diversity in \Cref{sec:neighboring,sec:convergence}.

Furthermore, the remaining baselines exhibit distinct failure modes. GCBC performs well on in-distribution goals (e.g., fixed-goal PushT) and in environments with dense data coverage (Medium Maze, Wall) but degrades significantly on out-of-distribution (OOD) goals. HIQL shows potential for OOD generalization but demands dense data coverage to accurately learn value functions, reflected in its performance gap between PushT and PushT-Aug. Diffuser-based methods struggle with dataset noise; despite adding replanning (\Cref{appendix:baseline_tuning_diffuser}), performance remains poor. While H-Diffuser offers marginal improvements, it still relies heavily on dense data. Even with data augmentation (\Cref{app:data_aug}), H-Diffuser's success rate remains low, primarily because it struggles to jointly attend to both the object and the agent's final position.

\textbf{Comparison with hierarchical latent world models.} We additionally compare with HWM~\citep{Zhang2026HierarchicalPW}, a hierarchical world model that increases the frameskip of its high-level model and compresses action sequences into macro-actions with a transformer-based action encoder. We compare with the DINO-WM variant of HWM. As shown in \Cref{tab:hwm}, on PushT, which requires long-horizon object manipulation, \method{} outperforms HWM by a large margin on all variants. This highlights the benefit of disentangling state reachability from fine-grained control using diffusion noise as an implicit latent action.

\begin{table}[h]
\centering
\caption{Success rate of \method{} and HWM on PushT environments with fixed (Fix) and random (Rand.) goals.}
\label{tab:hwm}
\vspace{0.5em}
\small
\begin{tabular}{@{}lcccc@{}}
\toprule
& \multicolumn{2}{c}{\textbf{PushT}} & \multicolumn{2}{c}{\textbf{PushT-Aug}} \\
\cmidrule(lr){2-3} \cmidrule(lr){4-5}
\textbf{Method} & Fix & Rand. & Fix & Rand. \\
\midrule
HWM & 0.38 & 0.20 & 0.42 & 0.16 \\
\method{} & \textbf{1.00} & \textbf{0.77} & \textbf{1.00} & \textbf{1.00} \\
\bottomrule
\end{tabular}
\end{table}

\begin{figure}[t!]
    \centering
    \includegraphics[width=0.8\columnwidth]{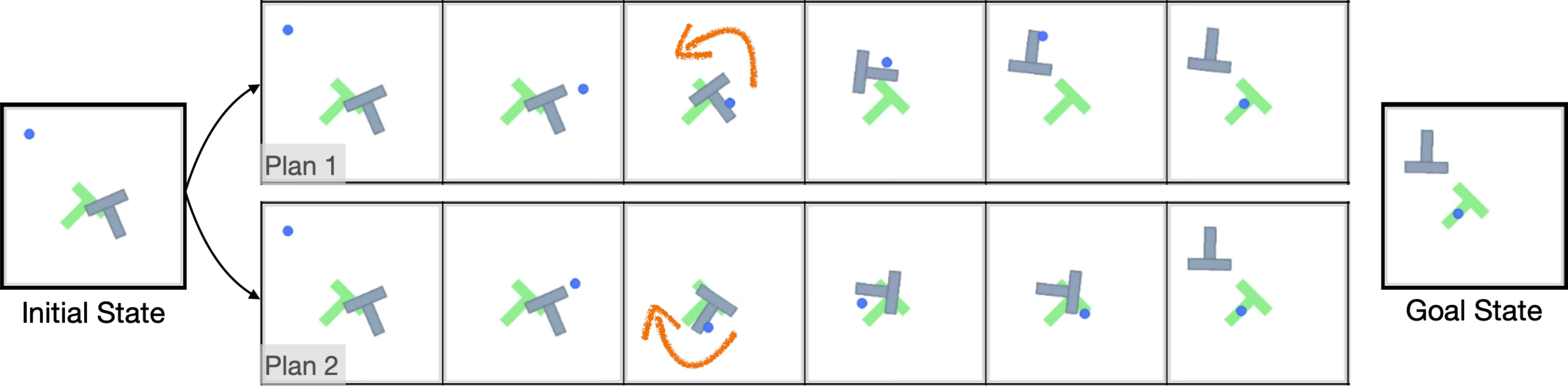}
    \caption{Diverse plans generated by \method{}. The top and bottom rows illustrate distinct strategies: counterclockwise and clockwise rotations of the T-block, both resulting in successful task completion. }
    \label{fig:diverse_plans}
\end{figure}

Unlike approaches that explicitly model dataset behavior, \method{} is not constrained by the underlying training distribution. By leveraging full planning over the environment dynamics, our method is inherently capable of finding diverse solutions for a given goal. \Cref{fig:diverse_plans} demonstrates this by showing two different subgoal sequences generated by \method{}. For instance, Plan 1 (top row) rotates the T-block counterclockwise, while Plan 2 (bottom row) rotates it clockwise. Ultimately, both planned trajectories successfully achieve the final objective via our hierarchical framework.

\subsection{How does \method{} scale with task complexity?}

\begin{figure}[t!]
    \centering
    \includegraphics[width=\columnwidth]{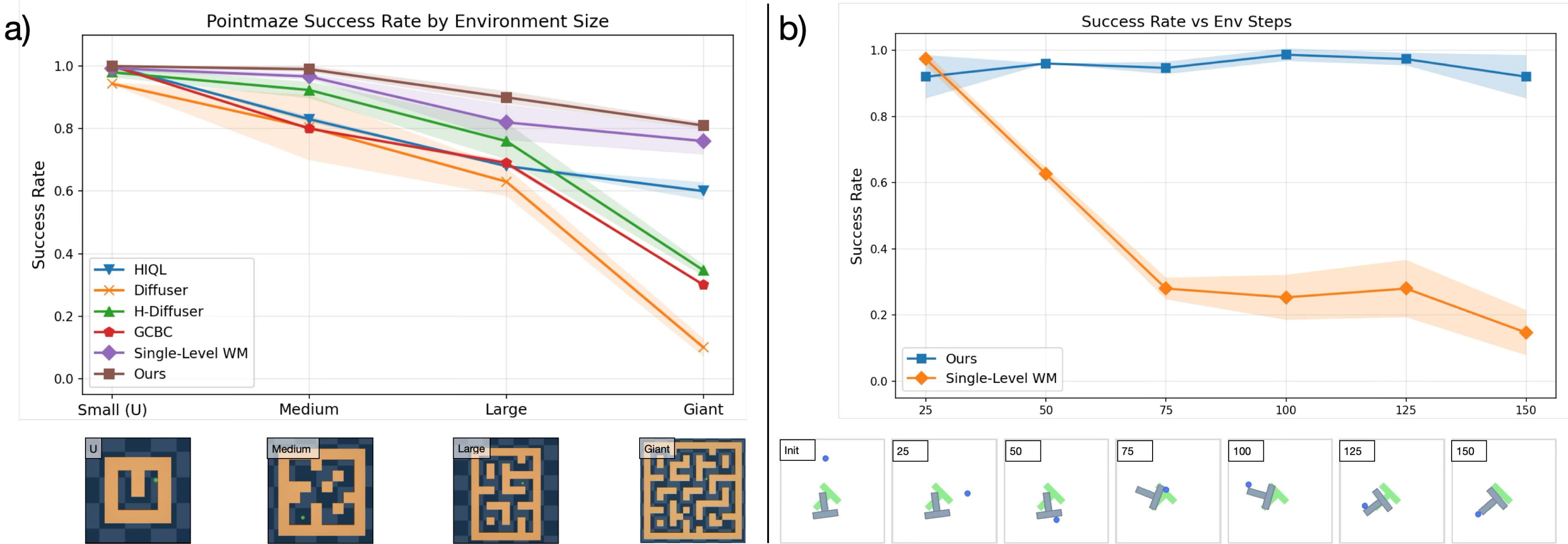}
    \caption{Planning performance with increasing task complexity. In a), we test \method{}'s ability on Maze with increasing sizes. In b), we test \method{} on increasingly longer horizon PushT tasks.}
    \label{fig:complexity}
\end{figure}

We investigate whether \method{} scales to tasks of increasing complexity across two dimensions: environment size and goal horizon.

\textbf{Scaling with Maze Size.} We evaluate \method{} across four mazes of increasing difficulty: U-Maze ($3 \times 3$), Medium Maze ($6 \times 6$), Large Maze ($7 \times 10$), and Giant Maze ($13 \times 13$) (\Cref{fig:complexity}a). Each dataset contains 2,000 random 100-step trajectories. Given a fixed dataset size, data coverage naturally becomes sparser as maze size increases.

While all methods achieve near-perfect success on the U-Maze, baselines degrade significantly in larger environments (\Cref{fig:complexity}a). In contrast, \method{} maintains the highest performance on the Large Maze. We hypothesize this is because \method{} combines local dynamics modeling with high-level planning, effectively "stitching" together regions of sparse data. Unlike GCBC or Diffuser (which model entire trajectories simultaneously) or HIQL (which requires dense data for global value functions), \method{} does not rely on global trajectory coverage. Furthermore, while H-Diffuser uses abstraction, it lacks an explicit dynamics model, resulting in brittle plans requiring expensive, high-frequency replanning. By grounding its hierarchy in explicit local dynamics, \method{} generates robust, scalable plans.

\textbf{Scaling with Goal Horizon.} Building on the PushT planning benchmark introduced by \citet{dinowm} (which evaluates goals 25 steps away), we increase complexity by sampling goals from the validation set that are $d \in \{25, 50, \dots, 150\}$ steps away.

We compare \method{} against a single-level world model baseline that performs full-horizon CEM search and replanning at every step. While the total number of executed actions is identical, \method{} achieves a near-perfect success rate across all horizons as shown in~\Cref{fig:complexity}b). This confirms our main PushT results (\Cref{tab:results}), demonstrating that \method{} maintains precise goal-reaching capabilities even when goals are over 150 steps away.

\subsection{Vision-based Planning with~\method{}}

To evaluate \method{} in high-dimensional observation spaces, we test it on a vision-based PushT task. We use the state-based PushT dataset augmented with four 90$^\circ$ rotations for spatial robustness. Observations ($3 \times 64 \times 64$ images) are encoded into $4 \times 8 \times 8$ latents using a pretrained Stable Diffusion VAE~\citep{stablediffusion}. We compare against AVDC~\citep{avdc}, a language-conditioned video planner that generates all subgoals simultaneously (details in \Cref{appendix:baselines}).

As shown in \Cref{fig:visual}, \method{} generates reliable, sequential subgoals that ensure steady progress. In contrast, AVDC fails to produce meaningful plans, likely because the initial-goal pairs are out-of-distribution. This highlights the generalization advantage of iterative subgoal search over simultaneous generation---which relies heavily on trajectory-level dataset coverage---even within compressed visual spaces.

\begin{figure}[H]
    \centering
    \includegraphics[width=0.7\columnwidth]{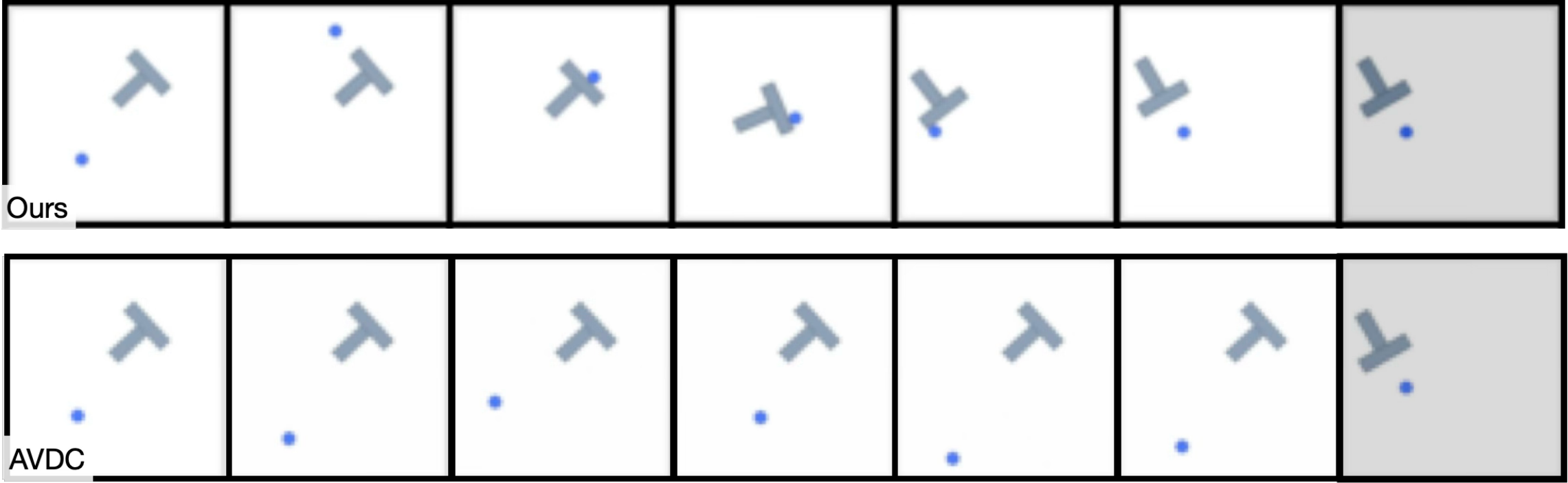}
    \caption{Subgoals generated by \method{} on a visual PushT task. The first column denotes the starting state, and the last shaded column denotes the desired goal state. \method{} generates reliable, sequential subgoals that drive steady progress toward the goal state without any degenerate states. }
    \label{fig:visual}
\end{figure} 

\textbf{Quantitative evaluation.} To complement the qualitative comparison, we report success rates on vision-based PushT in \Cref{tab:vision_quant}. We compare against two additional baselines: DINO-WM~\citep{dinowm}, a flat single-level world model planner, and HWM~\citep{Zhang2026HierarchicalPW}, a concurrent hierarchical latent world model that uses a larger frameskip for its high-level model. In this setting, \method{} uses DINO-WM as its low-level world model, so the comparison between DINO-WM and \method{} isolates the gain provided by the hierarchy. We evaluate four settings: for $d \in \{25, 50, 75\}$, the goal is sampled from a held-out validation set and is known to be $d$ steps away; for Random Goal, both the initial and goal states are sampled randomly from the environment. \method{} achieves the best performance on the most challenging long-horizon settings ($d = 50$, $d = 75$, and Random Goal), demonstrating its effectiveness for long-horizon vision-based planning.

\begin{table}[h]
\centering
\caption{Success rate on vision-based PushT. For $d \in \{25, 50, 75\}$ the goal is $d$ steps away; Random Goal samples both the initial and goal states randomly.}
\label{tab:vision_quant}
\vspace{0.5em}
\small
\begin{tabular}{@{}lcccc@{}}
\toprule
\textbf{Method} & $d = 25$ & $d = 50$ & $d = 75$ & \textbf{Random Goal} \\
\midrule
DINO-WM & 0.84 & 0.16 & 0.00 & 0.00 \\
HWM & \textbf{0.96} & 0.46 & 0.14 & 0.02 \\
\method{} & 0.90 & \textbf{0.60} & \textbf{0.50} & \textbf{0.26} \\
\bottomrule
\end{tabular}
\end{table}

\section{Conclusion}

We present \method{}, a hierarchical world modeling and planning framework that decouples temporally extended subgoal generation from local reachability. By treating diffusion noise as implicit actions, we constrain the planning search space to a unit Gaussian, ensuring reliable subgoal generation. \method{} significantly outperforms state-of-the-art baselines and demonstrates high-quality vision-based subgoal generation. Ultimately, grounding hierarchical abstraction in explicit local dynamics represents a promising step toward robust, long-horizon decision-making.

\textbf{Limitations.} First, \method{}'s reliance on search-based optimization within a diffusion latent space incurs higher inference times than reactive policies or deterministic world models. Future work will explore amortized or gradient-based planning to mitigate this bottleneck. Second, while our hierarchy prevents planning complexity from scaling with trajectory length, it still scales with state-space dimensionality. Integrating compact, structured representation learning could further streamline this planning process for highly complex environments.

\begin{ack}

We would like to thank Amir Bar, Alfredo Canziani, Yilun Kuang, Hengkai Pan, Mike Rabbat, Ying Wang, and Wancong Zhang for their valuable discussion and feedback. 
This work was supported by AFOSR under grant FA95502310139, the Institute of Information \& Communications Technology Planning \& Evaluation (IITP) with a grant funded by the Ministry of Science and ICT (MSIT) of the Republic of Korea in connection with the Global AI Frontier Lab International Collaborative Research Agreement (No. RS-2024-00469482 \& RS-2024-00509279).

\end{ack}

\newpage
{\small
\bibliographystyle{unsrtnat}
\bibliography{main}
}
\clearpage

\newpage
\section{Appendix}

\subsection{Environments and Dataset Generation}
\label{appendix:envs}
We evaluate~\method{} on three navigation and manipulation environment suites as shown in Figure~\ref{fig:envs}. The environments are Wall, PushT, and Maze, with task horizon ranging from hundreds to thousands of environment steps. 

\begin{enumerate}[label=\alph*), leftmargin=*]
    \item \textbf{Wall:} Introduced in~\citet{dinowm}, this is a simple 2D navigation environment that consists of two rooms separated by a wall with a door. The agent's goal is to navigate from a random starting location on one side of the wall to a random location on the other side, requiring the agent to pass through the door. We use the same dataset as in~\citet{dinowm}, which consists of randomly generated 1920 trajectories of length 50. 
    
    \item \textbf{Maze: } This environment is introduced in~\citet{d4rl}, where a force-actuated 2-DoF ball navigates to an arbitrary location in a maze. In this work, we consider maze layout of four sizes, U ($3 \times 3$), Medium ($6 \times 6$), Large ($7 \times 10$), and Giant ($13 \times 13$), visualized in Figure~\ref{fig:complexity} a). We generate 2000 random trajectories of length 100 for each maze layout.
    
    \item \textbf{PushT: } Introduced by \citet{diffusionpolicy}, this environment consists of a pusher agent and a T-shaped block. The objective is for both the agent and the block to reach specified target locations from random initial positions. We evaluate two variants of this task: in \emph{Fix Goal}, the target state is a fixed configuration (the green T); in \emph{Random Goal}, the target is randomly sampled from the environment's state space. Both settings are significantly more challenging than the formulation in~\citet{dinowm}, where the goal is provided at a known, fixed horizon of 25 steps. In contrast, completing the task in our settings often requires hundreds of steps, rendering standard single-level planning infeasible. Following \cite{dinowm}, we train on noisy, replayed trajectories. We further evaluate our method in a data-abundant regime by applying random translation and rotation augmentations to the existing dataset.

\end{enumerate}

\subsection{Task Visualizations}
\label{sec:task_define}

\begin{figure}[!t]
    \centering
    \includegraphics[width=0.9\linewidth]{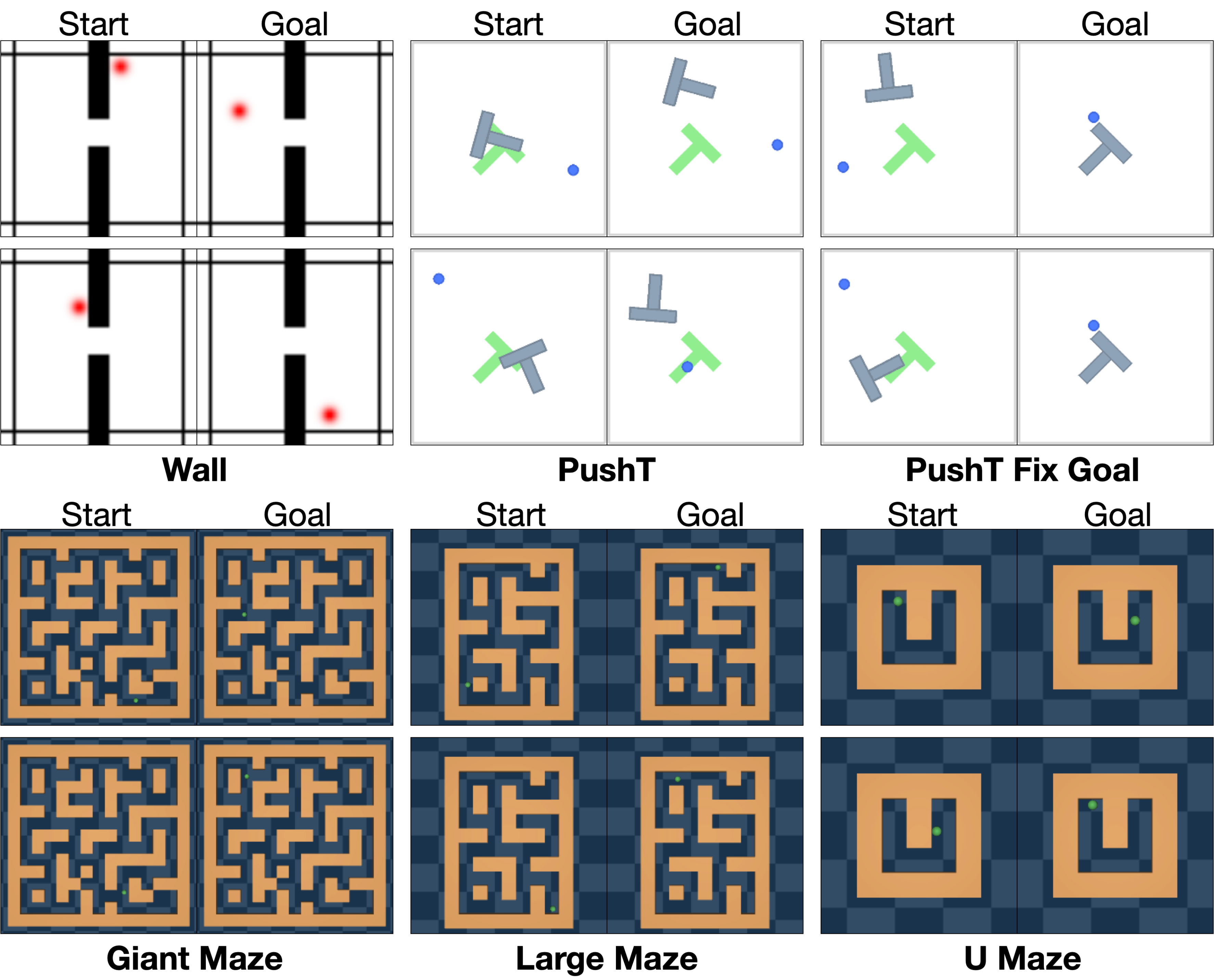}
    \caption{Examples of sampled initial and goal state pairs for the Wall, PushT, Giant Maze, Large Maze, and U-Maze environments}
    \label{fig:init_goal}
\end{figure}

For the Wall, PushT, Giant Maze, Large Maze, Medium Maze, and U-Maze environments, we randomly sample 50 initial and goal states from each environment’s state space (visualizations in \Cref{fig:init_goal}). For PushT, the random-goal setting samples both the initial and goal states randomly from the environment, whereas the fixed-goal setting samples a random initial state while using the environment’s fixed goal location. For the Wall environment, to ensure sufficient task difficulty, the initial and final object locations are constrained to lie on opposite sides of the wall. For the Giant Maze, Large Maze, Medium Maze, and U-Maze environments, we follow the random state sampling procedure used in OGBench: we sample arbitrary free cells for the ball and add continuous uniform noise around the corresponding grid-cell centers.

\subsection{Data Augmentation}
\label{app:data_aug}
\paragraph{Data augmentation for PushT.}
  For PushT we additionally apply geometric data augmentation to the offline dataset at each training step.
  The state vector contains end-effector position and velocity $(x_\mathrm{dot}, y_\mathrm{dot},
  \dot{x}_\mathrm{dot}, \dot{y}_\mathrm{dot})$, T-block position $(x_T, y_T)$, and T-block orientation
  $(\sin\theta_T, \cos\theta_T)$.
  One of three augmentation types is sampled per step:
  \begin{itemize}
      \item \textbf{Displacement} (40\%): all positions shifted by a random integer offset
            $(\delta x, \delta y)$ chosen so that both the end-effector and the T-block remain within
            the screen boundaries $[50,450]$ and $[100,400]$ respectively.
      \item \textbf{Rotation} (30\%): all positions and the T-block angle rotated by
            $\theta \sim \mathcal{U}(0, 2\pi)$ about the scene center; velocities and actions are
            rotated accordingly.
      \item \textbf{Displacement + Rotation} (30\%): a half-range displacement combined with a small
            rotation $\theta \sim \mathcal{U}(0, 30^\circ)$.
  \end{itemize}

\subsection{Baselines}
\label{appendix:baselines}

We compare \method{} with five state-of-the-art offline learning methods including imitation learning (GCBC), offline RL (HIQL), diffusion planners (Diffuser, Hierarchical Diffuser), and planning with world models (Single-level WM with MPC). For qualitative vision-based experiments, we compare \method{} with AVDC~\citep{avdc}.

\begin{enumerate}[label=\alph*), leftmargin=*]
    \item \textbf{Single-level WM and Planning:} This baseline employs a flat world model based on \citet{dinowm} to perform single-level planning via the Cross-Entropy Method (CEM) without hierarchical abstractions. 
    
    \item \textbf{Goal-conditioned Behavior Cloning (GCBC): } An imitation learning paradigm where a policy is trained to map the current state and a target goal observation directly to the expert action that leads toward that goal. For our implementation, we extend Diffusion Policy \citep{diffusionpolicy} by conditioning the policy on a future frame randomly sampled from the trajectory. 
    
    \item \textbf{HIQL~\citep{hiql}:} An offline reinforcement learning method that learns hierarchical representations by combining high-level subgoal-as-action policy with low-level subgoal-conditioned policy for long-horizon tasks.
    
    \item \textbf{Diffuser~\citep{diffuser}: } A trajectory-level diffusion model that performs planning by sampling entire state-action sequences. We replan after executing each action in the environment. 
    
    \item \textbf{Hierarchical Diffuser~\citep{hdiffuser}:} An extension of Diffuser that decomposes the planning process into two temporal scales. A high-level module synthesizes intermediate subgoals, which are subsequently used by a low-level denoising network to generate a refined trajectory. We replan both the high level and the low level after executing each action in the environment.
    
    \item \textbf{AVDC~\citep{avdc}:} This approach uses a video generation model to predict future states and then employs inverse kinematics (IK) to derive the necessary actions; we specifically use it to compare the quality of its generated video sequences against the high-level subgoals produced by our method.
    \item \textbf{HWM~\citep{Zhang2026HierarchicalPW}:} A concurrent hierarchical latent world model that increases the frameskip of the high-level model and plans over learned macro-actions. We compare with the DINO-WM variant of HWM. Since the official implementation of this variant is not open-sourced, we reimplement it following the reported settings and DINO-WM for unspecified details. We compare against HWM in \Cref{tab:hwm,tab:vision_quant}.
\end{enumerate}

\subsection{Hierarchical Planning with CEM}

In this section, we provide the complete pseudo-code for our hierarchical planning procedure using CEM. \Cref{alg:verbose} expands upon the simplified overview presented in~\Cref{alg:reduced}, explicitly detailing the sampling and optimization loops for both high-level subgoal generation and low-level action execution.

\begin{algorithm}[t]
\caption{Hierarchical Planning with \method{} (CEM)}
\label{alg:verbose}
\begin{algorithmic}[1]
\REQUIRE Initial state $s_0$, goal state $s_g$, high/low models $f_{hi}, f_{lo}$, costs $C_{hi}, C_{lo}$
\REQUIRE CEM parameters: $N$ samples, $M$ iterations, $K$ elites, horizons $H_{hi}, H_{lo}$
\STATE $s_t \gets s_0$
\WHILE{goal $s_g$ not reached}
    \STATE \textcolor{teal}{\ttfamily{\# High-Level Plan (Subgoal Generation)}}
    \STATE Initialize latent distribution parameters $\theta_{hi}$
    \FOR{$m = 1$ to $M$}
        \STATE Sample $N$ latent sequences $Z^{(n)} = (z^{(n)}_1, \dots, z^{(n)}_{H_{hi}}) \sim \mathcal{D}(\theta_{hi})$
        \FOR{$n = 1$ to $N$}
            \STATE Rollout subgoals: $\hat{s}^{(n)}_t \sim f_{hi}(\hat{s}^{(n)}_{t-1}, z^{(n)}_t)$ for $t \in [1, H_{hi}]$
            \STATE Compute trajectory cost $J^{(n)}_{hi} = C_{hi}(\hat{s}^{(n)}_{1:H_{hi}}, s_g)$
        \ENDFOR
        \STATE Update $\theta_{hi}$ using top $K$ elite sequences with the lowest $J_{hi}$
    \ENDFOR
    \STATE Extract best subgoal sequence $(\hat{s}^*_1, \dots, \hat{s}^*_{H_{hi}})$
    
    \STATE \textcolor{teal}{\ttfamily{\# Low-Level Plan (Action Generation)}}
    \FOR{$i = 1$ to $H_{hi}$}
        \STATE Initialize action distribution parameters $\theta_{lo}$
        \FOR{$m = 1$ to $M$}
            \STATE Sample $N$ action sequences $A^{(n)} = (a^{(n)}_1, \dots, a^{(n)}_{H_{lo}}) \sim \mathcal{D}(\theta_{lo})$
            \FOR{$n = 1$ to $N$}
                \STATE Rollout states: $\tilde{s}^{(n)}_t \sim f_{lo}(\tilde{s}^{(n)}_{t-1}, a^{(n)}_t)$ for $t \in [1, H_{lo}]$
                \STATE Compute trajectory cost $J^{(n)}_{lo} = C_{lo}(\tilde{s}^{(n)}_{1:H_{lo}}, \hat{s}^*_i)$
            \ENDFOR
            \STATE Update $\theta_{lo}$ using top $K$ elite sequences with the lowest $J_{lo}$
        \ENDFOR
        \STATE Extract best action sequence $(a^*_1, \dots, a^*_{H_{lo}})$
        \STATE \textcolor{teal}{\ttfamily{\# Execute Planned Actions}}
        \STATE Execute actions $(a^*_1, \dots, a^*_{H_{lo}})$ in the environment
        \STATE $s_t \gets$ observe new state
    \ENDFOR
\ENDWHILE
\end{algorithmic}
\end{algorithm}

\subsection{Ablation: Effect of CEM Search Budget and Noise Regularization}
\label{appendix:ab_plan_param}

\begin{figure}[t]
    \centering
    \includegraphics[width=0.8\columnwidth]{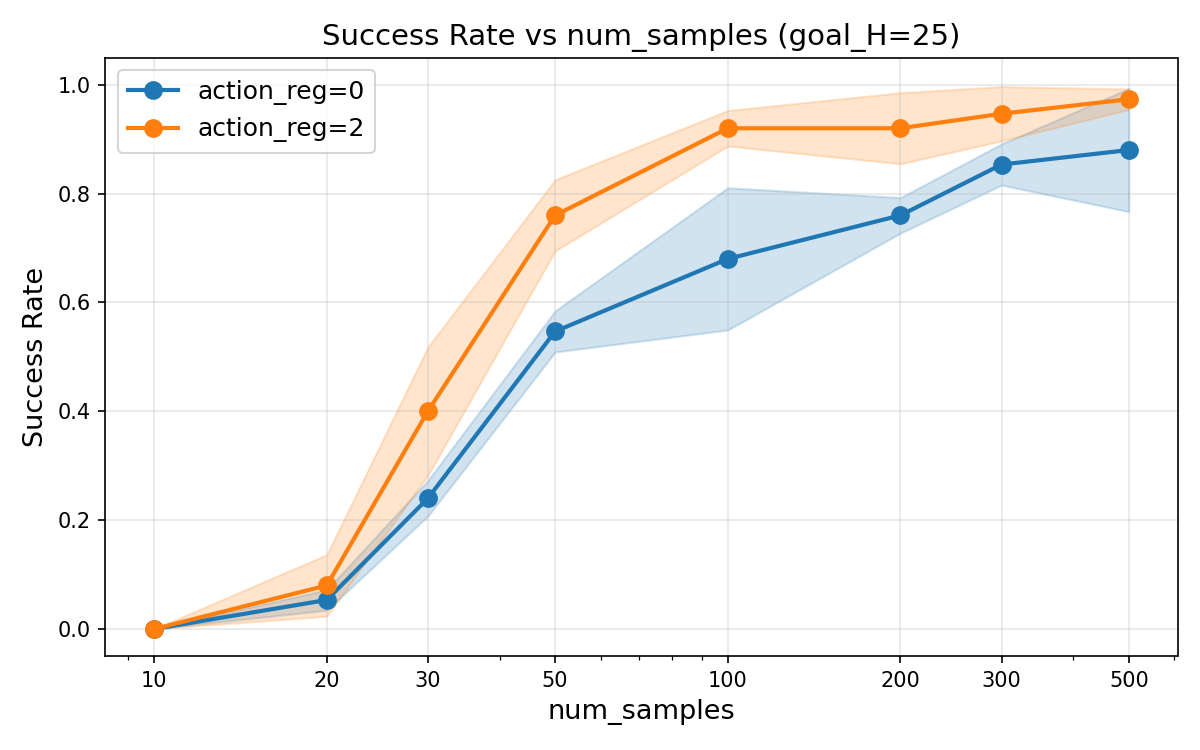}
    \caption{Success rate with respect to CEM samples and noise regularization.}
    \label{fig:cem_noise}
\end{figure} 

We investigate the impact of the CEM search budget and the noise regularization strategy introduced in \Cref{sec:planning}. By varying the number of CEM samples per level (maintaining a 10\% elite fraction), we demonstrate that the performance of \method{} scales effectively with test-time compute (\Cref{fig:cem_noise}). Furthermore, planning with non-zero noise regularization ($\alpha > 0$) consistently yields higher success rates compared to the unregularized baseline. Unless otherwise noted, all reported results use the same planner configuration at both levels: 300 CEM samples, 30 elites, and a horizon of 5 (\Cref{tab:cem_cfg}). 

\subsection{Ablation on $\Delta$ and number of subgoals}
\label{appendix:delta_ablation}
All results in the main paper use $\Delta = 5$ (25 environment steps with frameskip) and $H_{hi} = 5$ subgoals per high-level plan, without per-task tuning. State-based experiments use a 50-step noise schedule for both DDPM training and DDIM inference; vision-based experiments train with 250 steps and use 25 DDIM steps at inference.

To assess sensitivity to these choices, we sweep $\Delta$ and the number of subgoals on PushT-Aug with random goals (\Cref{tab:delta_ablation}). Performance is robust for large $\Delta$ across a wide range of subgoal counts. For a fixed effective search horizon ($\Delta \times$ number of subgoals), splitting it evenly between the two levels is most effective: at an effective horizon of roughly 25, the success rate increases monotonically from 0.06 ($\Delta = 1$, 25 subgoals) to 0.96 ($\Delta = 5$, 5 subgoals). A very small $\Delta$ pushes the burden onto a long subgoal chain while contributing little per hop.

\begin{table}[h]
\centering
\caption{Success rate on PushT-Aug (Random Goal) when sweeping the subgoal horizon $\Delta$ (rows) and the number of subgoals per high-level plan (columns). The default configuration ($\Delta = 5$, 5 subgoals) is underlined.}
\label{tab:delta_ablation}
\vspace{0.5em}
\small
\begin{tabular}{@{}c ccccccc@{}}
\toprule
$\Delta$ \textbackslash\ \#Subgoals & 3 & 4 & 5 & 6 & 8 & 12 & 25 \\
\midrule
1 & 0.06 & 0.04 & 0.06 & 0.08 & 0.04 & 0.10 & 0.06 \\
2 & 0.14 & 0.14 & 0.28 & 0.50 & 0.64 & 0.48 & 0.24 \\
3 & 0.50 & 0.56 & 0.64 & 0.82 & 0.76 & 0.74 & 0.42 \\
4 & 0.60 & 0.76 & 0.86 & 0.86 & \textbf{1.00} & 0.90 & 0.44 \\
5 & 0.74 & 0.84 & \underline{0.96} & 0.98 & 0.94 & 0.76 & 0.38 \\
\bottomrule
\end{tabular}
\end{table}

\subsection{Steps-to-goal metric}
\label{appendix:steps_to_goal}
Success rate measures whether a goal is reached but not how efficiently. To quantify trajectory efficiency, we report the average number of environment steps to success for all methods on PushT-Aug (Random Goal) and Medium Maze in \Cref{tab:steps_to_goal}, computed over successful trajectories only (N/A if no trajectory succeeded). \method{} is the most efficient method on PushT-Aug and the most efficient among value-function-free methods on Medium Maze. HIQL's learned value function yields substantially shorter maze paths, reflecting a trade-off: the planning cost in \Cref{sec:planning} is the distance between the last imagined state and the goal, which optimizes for reaching the goal rather than for path efficiency. Within the same framework, a self-supervised value function could be substituted for this cost to improve path optimality.

\begin{table}[h]
\centering
\caption{Average steps to success over successful trajectories (lower is better). N/A indicates that no trajectory succeeded.}
\label{tab:steps_to_goal}
\vspace{0.5em}
\small
\begin{tabular}{@{}lcc@{}}
\toprule
\textbf{Method} & \textbf{PushT-Aug (Rand.)} & \textbf{Medium Maze} \\
\midrule
HIQL & 307.4 & \textbf{34.1} \\
GCBC & 1014.9 & 336.8 \\
Diffuser & N/A & 287.8 \\
H-Diffuser & 713.9 & 169.4 \\
SL-WM & 312.5 & 229.5 \\
\method{} & \textbf{234.7} & 163.9 \\
\bottomrule
\end{tabular}
\end{table}

\subsection{Test-time Compute}
\label{appendix:test_time_compute}
\Cref{tab:inference_time} reports the wall-clock inference time to reach one goal for each method in the state-based setting, measured on a single NVIDIA L40S GPU. GCBC and HIQL are reactive policies whose compute budget cannot be adapted at test time; GCBC is comparatively slow because of its Diffusion Policy backbone. Diffuser and H-Diffuser follow their original formulation of replanning a full trajectory after every action, which is the most expensive setting. Despite running CEM at both levels, \method{} is roughly $27\times$ faster than Diffuser and $54\times$ faster than H-Diffuser, and faster than GCBC. The only faster methods are HIQL and SL-WM, both of which fail on long-horizon tasks.

The CEM search budget for \method{} and SL-WM is identical in all reported results (\Cref{tab:cem_cfg}): a horizon of 5 at each of our two levels gives an effective horizon of 25, matching SL-WM's flat horizon. \method{} is slower only because of DDIM sampling in the high-level model.

\textbf{Inference-time-matched SL-WM.} Since SL-WM runs faster than \method{} at the same CEM budget, we additionally match wall-clock time on PushT (Random Goal) by allowing SL-WM 45 replans ($3\times$). As shown in \Cref{tab:time_matched}, tripling SL-WM's replanning budget yields only a marginal improvement (0.21 to 0.28) and remains far below \method{} (0.77). This indicates that SL-WM's failure mode is compounding prediction error over long horizons rather than insufficient search, and that the gain comes from the hierarchy rather than from additional test-time compute.

\begin{table}[h]
\centering
\begin{minipage}[t]{0.42\textwidth}
\centering
\caption{Inference time to reach one goal on a single NVIDIA L40S GPU.}
\label{tab:inference_time}
\vspace{0.5em}
\small
\begin{tabular}{@{}lr@{}}
\toprule
\textbf{Method} & \textbf{Inference time} \\
\midrule
HIQL & 5.61 s \\
SL-WM & 16.93 s \\
\method{} & 53.08 s \\
GCBC & 73.3 s \\
Diffuser & 24 min \\
H-Diffuser & 48 min \\
\bottomrule
\end{tabular}
\end{minipage}
\hfill
\begin{minipage}[t]{0.54\textwidth}
\centering
\caption{CEM planner configuration shared by \method{} and SL-WM in all reported results.}
\label{tab:cem_cfg}
\vspace{0.5em}
\small
\begin{tabular}{@{}lcccc@{}}
\toprule
\textbf{Method} & \textbf{Samples} & \textbf{Elites} & \textbf{Replans} & \textbf{Horizon} \\
\midrule
SL-WM & 300 & 30 & 15 & 25 \\
\method{} & 300 & 30 & 15 & 5 (each level) \\
\bottomrule
\end{tabular}
\end{minipage}
\end{table}

\begin{table}[h]
\centering
\caption{Success rate on PushT (Random Goal) when matching either the CEM budget or the wall-clock inference time of SL-WM to \method{}.}
\label{tab:time_matched}
\vspace{0.5em}
\small
\begin{tabular}{@{}lccccc@{}}
\toprule
\textbf{Method} & \textbf{Samples} & \textbf{Elites} & \textbf{Replans} & \textbf{Horizon} & \textbf{SR} \\
\midrule
SL-WM (CEM budget matched) & 300 & 30 & 15 & 25 & 0.21 \\
SL-WM (inference time matched) & 300 & 30 & 45 & 25 & 0.28 \\
\method{} & 300 & 30 & 15 & 5 (each level) & \textbf{0.77} \\
\bottomrule
\end{tabular}
\end{table}

\subsection{Compute Resources}
\label{appendix:compute}

We train our state-based world models using 1 to 8 NVIDIA H200 Tensor Core GPUs, utilizing Distributed Data Parallel (DDP) to accelerate training when scaling to 8 GPUs. All baseline models are trained to convergence on a single NVIDIA H200 Tensor Core GPU. Planning evaluations are also conducted using a single NVIDIA H200 Tensor Core GPU. For the vision-based experiments, we employ a CDiT~\citep{nwm} backbone and train the model using DDP across a cluster of 32 NVIDIA H200 Tensor Core GPUs.

\subsection{Implementation Details}

\subsubsection{Implementations and Hyperparameters}
\label{sec:impl_hyperparams}
Hyperparameters for \method{} high-level dynamics model training and planning are listed in Table \ref{tab:hl_train_hyperparams} and Table \ref{tab:plan_hyperparams}, respectively. Hyperparameters for low-level WM training are listed in Table \ref{tab:wm_env_hyperparams} and Table \ref{tab:wm_shared_hyperparams}, while those for low-level WM planning are also included in Table \ref{tab:plan_hyperparams}. The high-level dynamics model and low-level WM share the same CEM planner configurations, except that action regularization is applied only in high-level planning. We use consistent hyperparameters across tasks, except for the observation history length in the Wall environment, which follows DINO-WM.

For success metrics, we continue using all state-based evaluation metrics from DINO-WM. Success is evaluated, and the planner may stop, after each plan is fully executed.

\begin{table*}[!t]
\centering
\small
\begin{minipage}[t]{0.49\textwidth}
\caption{Shared hyperparameters for training high-level dynamics model of \method{} across all environments.}
\label{tab:hl_train_hyperparams}
\vspace{2pt}
\centering
\begin{tabular}{@{}lc@{}}
\toprule
Name & Value \\
\midrule
Emb Dim & 56 \\
Down Dims & [256] \\
Kernel & 3 \\
Groups & 8 \\
lr & $1\mathrm{e}{-4}$ \\
WDecay & 0.1 \\
Noise Timesteps & 50 \\
lr warmup steps & 200 \\
\bottomrule
\end{tabular}
\end{minipage}
\hfill
\begin{minipage}[t]{0.49\textwidth}
\caption{Shared hyperparameters for planning with \method{} across all environments.}
\label{tab:plan_hyperparams}
\vspace{2pt}
\centering
\begin{tabular}{@{}lc@{}}
\toprule
Name & Value \\
\midrule
Noise Reg. coefficient $\alpha$ & $\{0,2\}$ \\
$\tau$ & 2 \\
$p$ & 1 \\
Max Replan & 15 \\
HL Horizon & 5 \\
LL Horizon & 5 \\
Top-$k$ & 30 \\
Num Samples & 300 \\
Var Scale & 1 \\
Opt Steps & 30 \\
\bottomrule
\end{tabular}
\end{minipage}

\vspace{4pt}
\begin{minipage}{0.98\textwidth}
\small
$\alpha$ is the coefficient on the noise regularization term $\mathcal{L}_{reg}$ in the high-level planning cost (\Cref{sec:planning}); $\tau$ and $p$ are its threshold and exponent.
\end{minipage}
\vspace{2pt}
\end{table*}

\begin{table*}[t]
\centering
\small
\begin{minipage}[t]{0.49\textwidth}
\caption{Environment-dependent hyperparameters for WM training. $T_o$ denotes the observation history length.}
\label{tab:wm_env_hyperparams}
\vspace{2pt}
\centering
\resizebox{\linewidth}{!}{%
\begin{tabular}{@{}lcccc@{}}
\toprule
 & $T_o$ & Frameskip & Dataset Size & Traj. Len. \\
\midrule
PushT      & 3 & 5 & 18500 & 100--300 \\
Wall        & 1 & 5 & 1920  & 50 \\
U-Maze      & 3 & 5 & 2000  & 100 \\
Medium Maze & 3 & 5 & 2000  & 100 \\
Large Maze  & 3 & 5 & 2000  & 100 \\
Giant Maze  & 3 & 5 & 2000  & 100 \\
\bottomrule
\end{tabular}%
}
\end{minipage}
\hfill
\begin{minipage}[t]{0.49\textwidth}
\caption{Shared hyperparameters for WM training.}
\label{tab:wm_shared_hyperparams}
\vspace{2pt}
\centering
\begin{tabular}{@{}lc@{}}
\toprule
Name & Value \\
\midrule
Predictor lr        & $5\mathrm{e}{-4}$ \\
Action encoder lr   & $5\mathrm{e}{-4}$ \\
Depth               & 6 \\
Heads               & 8 \\
MLP dim             & 64 \\
Dropout             & 0 \\
Emb. dropout        & 0 \\
Pool                & Mean \\
\bottomrule
\end{tabular}
\end{minipage}
\vspace{2pt}
\end{table*}

\subsubsection{Baseline Implementations and Hyperparameters}
Unless otherwise stated, all baselines are trained on the same offline datasets and evaluated using the same task specifications, goal definitions, and environment interaction budget as \method{}.

\subsubsection{Single-level WM and Planning}
We implement a single-level WM with transformer-based predictor following DINO-WM \citep{dinowm}. The training and inference details are identical to those of the low level WM in Section~\ref{sec:low_level_wm} with implementation details in Section~\ref{sec:impl_hyperparams}. During inference, the planner optimizes planning cost $\|\hat{s}_{H_{lo}} - s^*\|_2$. For state-based WM, we adapt DINO-WM and remove the frozen pre-trained DINOv2 encoder. This single-level WM baseline isolates the effect of hierarchy and tests whether subgoal proposal mitigates cumulative prediction error.

\subsubsection{Goal-conditioned Behavior Cloning (GCBC)}

GCBC trains a policy to imitate dataset actions conditioned on both the current state and a desired goal state. For training, given an offline trajectory $\{(s_t, a_t)\}_{t=0}^{T}$, we randomly sample a current state $s_t$ and a future state $s_{t+k}$ where $k>0$. The policy is trained to predict dataset action $a_t$ with the training objective:
\begin{equation}
\mathcal{L}_{\mathrm{GCBC}}=\mathbb{E}_{(s_t,a_t,g)\sim \mathcal{D}}
    \left[\| \pi_\theta(s_t, g) - a_t \|_2^2\right],
\end{equation}
Specifically, we implement GCBC as a Diffusion Policy~\citep{diffusionpolicy} conditioned on the concatenation of the current observation and goal state to predict an action chunk.

We include GCBC as a representative of direct goal-conditioned imitation learning and as a baseline for what behavioral cloning alone can achieve. Hyperparameters are listed in Table \ref{tab:gcbc_env_hyperparams} and Table \ref{tab:gcbc_shared_hyperparams}.

\begin{table*}[!t]
  \centering
  \small
  \begin{minipage}[t]{0.49\textwidth}
  \caption{Environment-dependent hyperparameters for GCBC training. $T_o$: observation history length; Act.\ Window: minimum timestep gap between current state and goal state; Pred.\ Horizon: action length generated per diffusion forward pass; Act.\ Steps: actions executed before re-planning.}
  \label{tab:gcbc_env_hyperparams}
  \vspace{2pt}
  \centering
  \resizebox{\linewidth}{!}{%
  \begin{tabular}{@{}lcccc@{}}
  \toprule
   & $T_o$ & Act. Window & Pred. Horizon & Act. Steps \\
  \midrule
  PushT      & 2 & 9  & 10 & 8  \\
  Wall        & 2 & 19 & 20 & 20 \\
  U-Maze      & 2 & 19 & 20 & 20 \\
  Medium Maze & 2 & 19 & 20 & 20 \\
  Large Maze  & 2 & 19 & 20 & 20 \\
  Giant Maze  & 2 & 19 & 20 & 20 \\
  \bottomrule
  \end{tabular}%
  }
  \end{minipage}
  \hfill
  \begin{minipage}[t]{0.49\textwidth}
  \caption{Shared hyperparameters for GCBC training.}
  \label{tab:gcbc_shared_hyperparams}
  \vspace{2pt}
  \centering
  \begin{tabular}{@{}lc@{}}
  \toprule
  Name & Value \\
  \midrule
  lr                  & $1\mathrm{e}{-4}$ \\
  Weight decay        & 0 \\
  Diffusion steps     & 100 \\
  Layers              & 8 \\
  Heads               & 4 \\
  Embedding dim       & 256 \\
  Attn.\ dropout      & 0.1 \\
  Emb.\ dropout       & 0 \\
  \bottomrule
  \end{tabular}
  \end{minipage}
  \vspace{2pt}
  \end{table*}

\subsubsection{HIQL~\citep{hiql}}

HIQL decomposes offline goal-conditioned RL into two levels of hierarchy. A high-level policy proposes a subgoal representation $\phi([s; w])$, where $w$ is an intermediate waypoint state sampled from the same trajectory, and a low-level policy executes primitive actions to reach that subgoal. Both policies are trained offline via Advantage-Weighted Regression (AWR) using a shared implicit value function $V_\theta(s, g)$.

The value function is trained with an IQL-style expectile regression objective:
\[ \mathcal{L}_V(\theta_V) = \mathbb{E}_{(s,s') \sim \mathcal{D},\, g \sim p(g \mid \tau)}
\left[ L_2^\tau \left( r(s,g) + \gamma V_{\bar{\theta}_V}(s',g) - V_{\theta_V}(s,g) \right) \right]. \]
where $L_2^\tau$ denotes the asymmetric squared expectile loss with parameter $\tau$.
Given the shared value function, both policies are updated via AWR. For the low-level policy $\pi^l$, the advantage measures progress toward subgoal $w$. For the high-level policy $\pi^h$, the advantage measures progress toward the final goal $g$ upon reaching subgoal $w$.
  
We use the HIQL implementation from the OGBench repository \citep{OGBench}. In all state-based experiments, no visual encoder is used, and $\phi$ processes raw state vectors directly. As HIQL is stochastic, all results are averaged over 3 random seeds. Hyperparameters are listed in Table \ref{tab:hiql_hyperparams}.

\begin{table}[t]
  \centering
  \small
  \caption{Shared hyperparameters for HIQL across all evaluation environments.}
  \label{tab:hiql_hyperparams}
  \begin{tabular}{@{}lc@{}}
  \toprule
  Name & Value \\
  \midrule
  lr & $3\mathrm{e}{-4}$ \\
  Discount factor $\gamma$ & 0.99 \\
  Expectile $\tau$ & 0.7 \\
  High-level $\alpha$ & 3.0 \\
  Low-level $\alpha$ & 3.0 \\
  Subgoal steps $k_{\mathrm{sub}}$ & 25 \\
  Actor/value hidden dims & $(512,512,512)$ \\
  Train steps & $10^7$ \\
  \bottomrule
  \end{tabular}
  \vspace{2pt}
\end{table}

\subsubsection{Diffuser~\citep{diffuser}}
Diffuser frames offline goal-conditioned planning as a trajectory denoising problem. A temporal U-Net processes a full trajectory $\tau = (s_0, a_0, \ldots, s_{H-1}, a_{H-1})$ of planning horizon $H$, and iterative reverse diffusion over $T$ steps is guided by goal conditioning to produce trajectories that reach a specified goal state. At inference, the current state is inpainted at the start of the trajectory and the goal state at the end; the remaining timesteps are filled by denoising. All environments share the same base configuration. Only the planning horizon $H$ is adjusted (in Table~\ref{tab:diffuser_env_hyperparams}).

Detailed implementation choices for inference are expanded in Section~\ref{appendix:baseline_tuning_diffuser}.

 \begin{table*}[t]
    \centering
    \small
    \begin{minipage}[t]{0.54\textwidth}
    \caption{Environment-dependent hyperparameters for Diffuser and HDiffuser.
             $H$: planning horizon; $J$: subsampling jump (HDiffuser high-level planner only).}
    \label{tab:diffuser_env_hyperparams}
    \vspace{2pt}
    \centering
    \begin{tabular}{@{}lcccc@{}}
    \toprule
     & Diffuser & \multicolumn{2}{c}{HDiffuser} \\
    \cmidrule(lr){2-2} \cmidrule(l){3-4}
     & $H$ & $H$ & $J$ \\
    \midrule
    PushT      & 128 & 125 & 25 \\
    Wall        & 48  & 45  & 15 \\
    U-Maze      & 48  & 75  & 25 \\
    Medium Maze & 48  & 75  & 25 \\
    Large Maze  & 48  & 75  & 25 \\
    Giant Maze  & 96  & 75  & 25 \\
    \bottomrule
    \end{tabular}%
    \end{minipage}
    \hfill
    \begin{minipage}[t]{0.43\textwidth}
    \caption{Shared hyperparameters for Diffuser and HDiffuser training.}
    \label{tab:diffuser_shared_hyperparams}
    \vspace{2pt}
    \centering
    \begin{tabular}{@{}lc@{}}
    \toprule
    Name & Value \\
    \midrule
    lr                   & $2\mathrm{e}{-4}$ \\
    Diffusion steps $T$  & 256 \\
    EMA decay            & 0.995 \\
    U-Net dim.\ mults    & $(1, 4, 8)$ \\
    \bottomrule
    \end{tabular}
    \end{minipage}
    \vspace{2pt}
  \end{table*}

\subsubsection{Hierarchical Diffuser~\citep{hdiffuser}}
H-Diffuser~\citep{hdiffuser} extends Diffuser with a two-level hierarchy. A high-level diffusion model plans jumpy subgoals spaced $J$ timesteps apart, producing a compressed sequence of length $\lceil H_{\mathrm{hl}}/J \rceil$ that covers a larger temporal horizon at lower computational cost. A \emph{low-level} diffusion model then plans a dense trajectory of length $H_{\mathrm{ll}}$ between consecutive subgoals, conditioned on the current state and the subgoal as the terminal constraint.

Hyperparameters shared with Diffuser are listed in Table~\ref{tab:diffuser_shared_hyperparams}. Environment-specific hyperparameters are listed in Table~\ref{tab:diffuser_env_hyperparams}.

Detailed implementation choices for inference are expanded in Section~\ref{appendix:baseline_tuning_diffuser}.

\subsubsection{Baseline Tuning for Diffuser \& H-Diffuser}
\label{appendix:baseline_tuning_diffuser}
The original Diffuser paper uses two execution strategies depending on the environment: for D4RL locomotion tasks, only the first action of each plan is executed and the model replans at every step (\textbf{execute one action + replan}); for maze2d and Kuka block-stacking, the full planned trajectory is executed open-loop without replanning (\textbf{execute plan}). H-Diffuser adopts the same two strategies at the subgoal level. To enable a fair comparison with our method, we additionally evaluate both models under a third strategy in which the full plan is executed and the model then replans (\textbf{execute plan + replan}).
Results under all three strategies are reported in Tables~\ref{tab:diffuser-pusht} and~\ref{tab:hdiffuser-pusht}.
    \begin{table*}[h]
    \centering
    \caption{Diffuser results across execution settings and PushT variants.}
    \label{tab:diffuser-pusht}
    \resizebox{\textwidth}{!}{%
    \begin{tabular}{@{}l l cc cc cc cc@{}}
    \toprule
    \multirow{3}{*}{\textbf{Model}} & \multirow{3}{*}{\textbf{Execution}} 
    & \multicolumn{4}{c}{\textbf{PushT}} 
    & \multicolumn{4}{c}{\textbf{PushT-Aug}} \\
    \cmidrule(lr){3-6} \cmidrule(l){7-10}
    & & \multicolumn{2}{c}{\textbf{Fix Goal}} & \multicolumn{2}{c}{\textbf{Random Goal}} 
    & \multicolumn{2}{c}{\textbf{Fix Goal}} & \multicolumn{2}{c}{\textbf{Random Goal}} \\
    \cmidrule(lr){3-4} \cmidrule(lr){5-6} \cmidrule(lr){7-8} \cmidrule(l){9-10}
    & & \textbf{SR} & \textbf{Cov} & \textbf{SR} & \textbf{Cov} & \textbf{SR} & \textbf{Cov} & \textbf{SR} & \textbf{Cov} \\
    \midrule
    Diffuser & execute plan + replan         & 0.00 & 0.03 & 0.00 & 0.03 & 0.28 & 0.37 & 0.06 & 0.19 \\
    Diffuser & execute one action + replan   & 0.00 & 0.11 & 0.00 & 0.06 & 0.04 & 0.27 & 0.02 & 0.15 \\
    Diffuser & execute plan                  & 0.00 & 0.05 & 0.00 & 0.03 & 0.00 & 0.09 & 0.00 & 0.13 \\
    \bottomrule
    \end{tabular}%
    }
    \end{table*}

     \begin{table*}[h]
    \centering
    \caption{H-Diffuser results across execution settings and PushT variants.}
    \label{tab:hdiffuser-pusht}
    \resizebox{\textwidth}{!}{%
    \begin{tabular}{@{}l l cc cc cc cc@{}}
    \toprule
    \multirow{3}{*}{\textbf{Model}} & \multirow{3}{*}{\textbf{Execution}} 
    & \multicolumn{4}{c}{\textbf{PushT}} 
    & \multicolumn{4}{c}{\textbf{PushT-Aug}} \\
    \cmidrule(lr){3-6} \cmidrule(l){7-10}
    & & \multicolumn{2}{c}{\textbf{Fix Goal}} & \multicolumn{2}{c}{\textbf{Random Goal}} 
    & \multicolumn{2}{c}{\textbf{Fix Goal}} & \multicolumn{2}{c}{\textbf{Random Goal}} \\
    \cmidrule(lr){3-4} \cmidrule(lr){5-6} \cmidrule(lr){7-8} \cmidrule(l){9-10}
    & & \textbf{SR} & \textbf{Cov} & \textbf{SR} & \textbf{Cov} & \textbf{SR} & \textbf{Cov} & \textbf{SR} & \textbf{Cov} \\
    \midrule
    H-Diffuser & execute plan + replan       & 0.38 & 0.28 & 0.00 & 0.08 & 0.56 & 0.51 & 0.46 & 0.36 \\
    H-Diffuser & execute one action + replan & 0.00 & 0.25 & 0.02 & 0.11 & 0.56 & 0.56 & 0.26 & 0.50 \\
    H-Diffuser & execute plan                & 0.04 & 0.09 & 0.00 & 0.06 & 0.02 & 0.17 & 0.00 & 0.10 \\
    \bottomrule
    \end{tabular}%
    }
    \end{table*}

\subsection{High-Level World Model Training Convergence}
\label{sec:convergence}

\begin{figure}[!t]
    \centering
    \includegraphics[width=\linewidth]{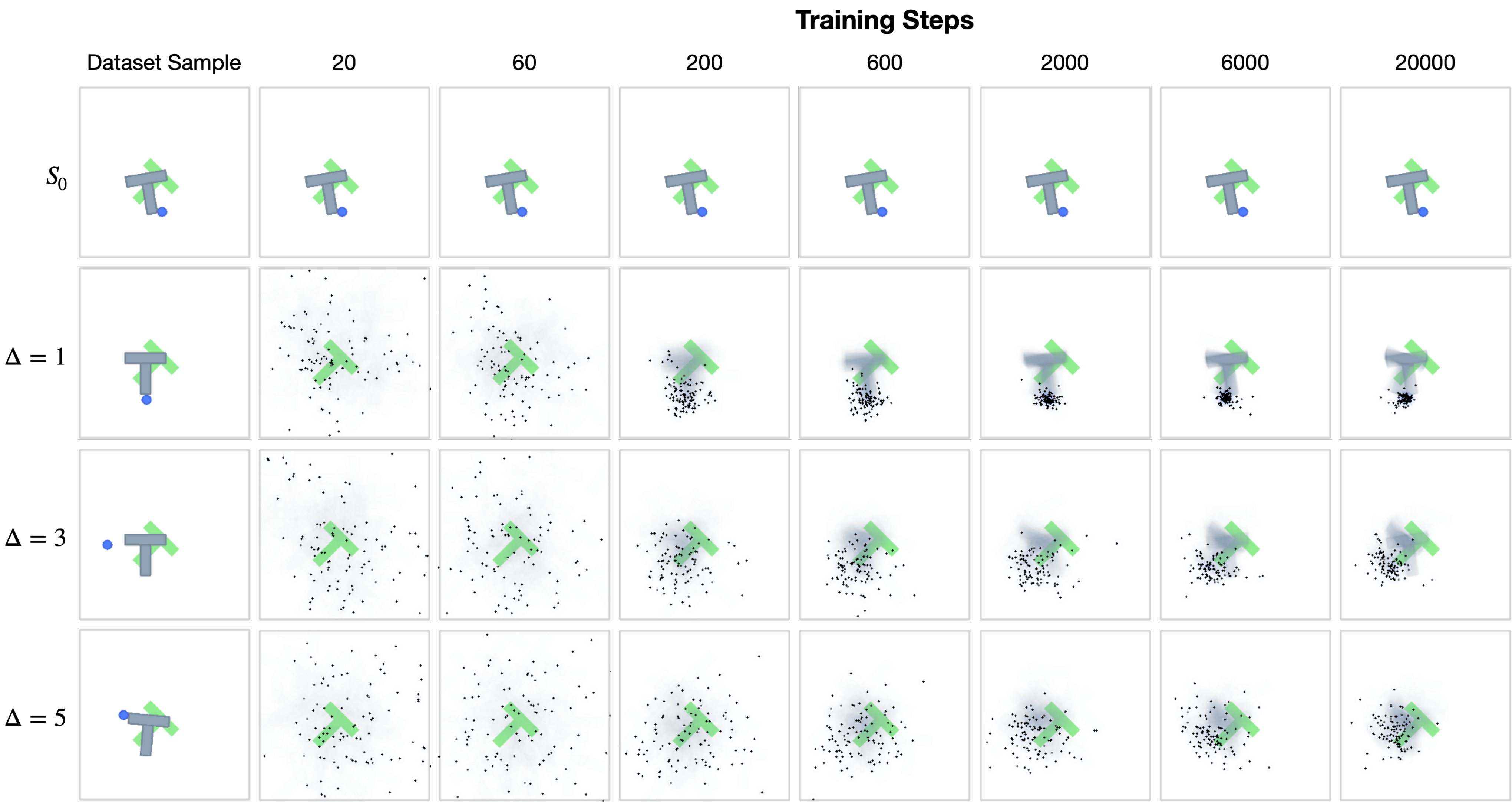}
    \caption{Training-time convergence of predicted future-state distributions for the PushT environment}
    \label{fig:training_distribution_convergence}
\end{figure}

We further analyze the training-time convergence behavior of the high-level world model, extending the qualitative study in \Cref{fig:next_states}. While \Cref{fig:next_states} visualizes the final learned neighboring-state distribution, \Cref{fig:training_distribution_convergence} shows how this distribution emerges over the course of training. 

Specifically, we fix validation observations $s_0$ and periodically sample predicted future states from intermediate checkpoints of the high-level WM. The first row shows the initial observation $s_0$ sampled from the validation set. The lower rows visualize generated future states $\hat{s}_\Delta = f_{hi}(s_0, z, \Delta)$ for horizons $\Delta \in \{1,3,5\}$, where each prediction is obtained by sampling latent noise vectors $z \sim \mathcal{N}(0,I)$. Each column corresponds to a different number of training steps on a logarithmic scale: 20, 60, 200, 600, 2000, 6000, and 20000.

As training begins, the predicted distributions are highly dispersed and often fail to preserve the structure of the PushT scene. In early checkpoints, the sampled futures do not consistently respect the local geometry of the object or the physically plausible displacement induced by the agent-object interaction. As training progresses, the generated samples become increasingly concentrated around coherent future configurations. By approximately 200 to 600 training steps, the model begins to capture the coarse reachable region for each temporal horizon, and later checkpoints further sharpen the predicted distribution while preserving meaningful sample diversity.

\subsection{Comparison of \method{} and naive WM with equivalent frameskip}
\label{app:vs}
\begin{figure}[!t]
    \centering
    \includegraphics[width=0.85\linewidth]{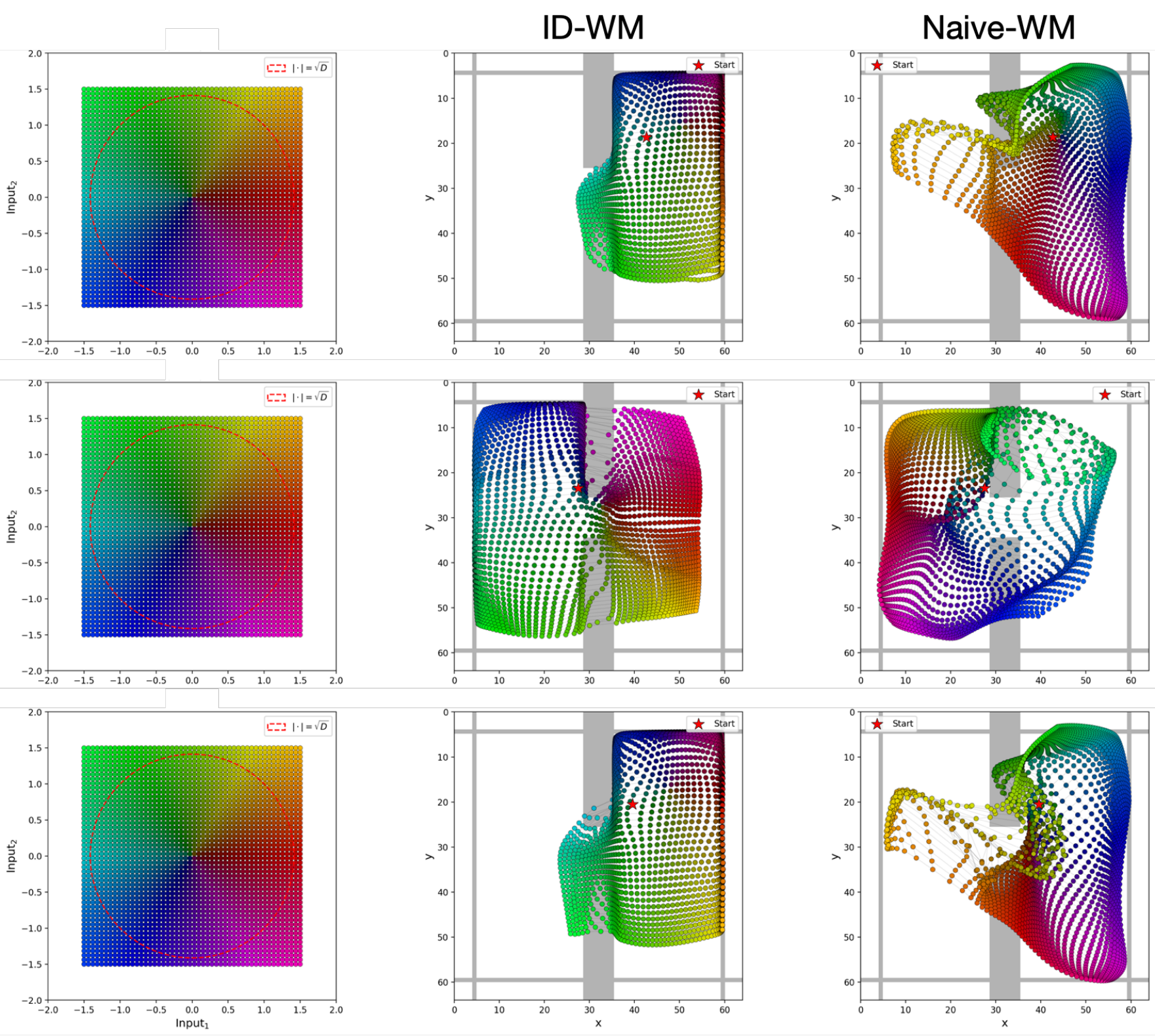}
    \caption{Visualization of the induced state mappings by \method{} and naive WM.}
    \label{fig:noise_mapping}
\end{figure}
We visualize the difference between our proposed \method{} high-level world model and a naive WM without implicit dynamics. The left column shows the sampled noise/action distribution. In \Cref{fig:noise_mapping}, the middle column shows the neighboring states predicted by \method{}, and the right column shows predictions from a naive WM with frameskip matched to the \method{} subgoal horizon. \method{} yields a smooth and locally consistent reachable set. The naive WM produces irregular mappings: nearby samples can jump to distant states, leave the local reachable neighborhood, or form discontinuous regions. This indicates that long-skip prediction with a naive world model is prone to out-of-distribution subgoals and discontinuous action-to-state geometry, making it less suitable for planning.

\end{document}